\documentclass[journal]{IEEEtran}

\usepackage{graphicx} 
\usepackage[caption=false]{subfig}
\usepackage{cite}
\usepackage{multirow}
\usepackage{enumitem}
\usepackage{booktabs}
\usepackage{amsmath,amssymb,amsfonts}
\usepackage{pifont}
\usepackage{algorithmic}
\usepackage[ruled,vlined,lined,linesnumbered,boxed,commentsnumbered]{algorithm2e}
\usepackage{booktabs}
\usepackage{url}
\usepackage{color}
\usepackage{balance}
\usepackage{makecell}
\usepackage{footmisc}
\usepackage{tikz}
\usetikzlibrary{calc}

\usepackage{./slam_macros}

\graphicspath{ {./images/}}

\begin{document}
\bstctlcite{IEEEexample:BSTcontrol}
%
\title{Good Graph to Optimize: Cost-Effective, Budget-Aware Bundle Adjustment in Visual SLAM}
%
%
%

\author{Yipu~Zhao,~\IEEEmembership{Member,~IEEE,}
Justin~S.~Smith,~\IEEEmembership{Student Member,~IEEE,}
Patricio~A.~Vela,~\IEEEmembership{Member,~IEEE,}
\thanks{Y.~Zhao is with Facebook Reality Lab (FRL).  
This work was done when Y.~Zhao was with the 
School of Electrical and Computer Engineering, 
Georgia Institute of Technology.  
J.~S.~Smith and P.~A.~Vela are with the 
School of Electrical and Computer Engineering,
and Institute of Robotics and Intelligent Machines,  
Georgia Institute of Technology.
E-mails: yipu.zhao@gatech.edu, jssmith@gatech.edu, pvela@gatech.edu.}
\thanks{This research was funded in part by the National Science 
Foundation (Award \#1816138).}
}


%

\maketitle

\begin{abstract}
The cost-efficiency of visual(-inertial) SLAM (VSLAM) is a critical
characteristic of resource-limited applications.
While hardware and algorithm advances have been significantly improved
the cost-efficiency of VSLAM front-ends, the cost-efficiency of VSLAM
back-ends remains a bottleneck.  
This paper describes a novel, rigorous method to improve the
cost-efficiency of local BA in a BA-based VSLAM back-end.  
An efficient algorithm, called {\em Good Graph}, is developed to select
size-reduced graphs optimized in local BA with condition preservation.
To better suit BA-based VSLAM back-ends, the {\em Good Graph} predicts
future estimation needs, dynamically assigns an appropriate size
budget, and selects a condition-maximized subgraph for BA estimation.
Evaluations are conducted on two scenarios: 
1) VSLAM as standalone process, and 
2) VSLAM as part of closed-loop navigation system. Results from the
first scenario show {\em Good Graph} improves accuracy and
robustness of VSLAM estimation, when computational limits exist. 
Results from the second scenario, indicate that {\em Good Graph} benefits the
trajectory tracking performance of VSLAM-based closed-loop navigation
systems, which is a primary application of VSLAM. 


\end{abstract}


%
\IEEEpeerreviewmaketitle


\section{Introduction} \label{sec::intro}

Visual(-inertial) simultaneous localization and mapping (VSLAM) has 
applications in robotics and augmented reality (AR) that span a diverse range
of platforms.  Consequently, the computational resources of robotics and
AR implementations are equally variable.
For instance, a micro flying vehicle requires lightweight computing
kits, and an AR headset typically has an ARM-based System-On-Chip (SoC) 
with low power consumption.  
At the other extreme, autonomous vehicles have more room and power for
compute hardware.
Though many state-of-the-art VSLAM systems achieve real-time performance
on a PC or a laptop with a powerful CPU, there may be underlying
processes hampered by the sequencing or timing of the pipeline
\cite{ye2019characterizing}. 
Some VSLAM systems fail to achieve real-time processing when subject 
to computational or time limits \cite{delmerico2018benchmark}.  
Others prioritize efficiency but exhibit performance loss 
\cite{SVO2017,sun2018robust}. 
The impacts and trade-offs magnify when attempting to replicate the same
level of performance on less powerful devices \cite{cadena2016past}.  
For practical robotics and AR applications, the cost-efficiency 
of VSLAM is essential and needs improvement.

Efficient VSLAM front-ends, that is, visual feature tracking, has been
pursued from both hardware and algorithmic perspectives.  
Dedicated processing units such as FPGAs
\cite{quigley2019open,liu2019eslam} and GPUs \cite{nagy2020faster} 
accelerate the highly parallelizable feature extraction step.  
Meanwhile, better algorithm designs, such as the incorporation of active
feature matching \cite{zhao2020gfm}, improve front-end runtime properties. 
While this progress enable cost-effective VSLAM front-ends, the
cost-efficiency of the VSLAM back-end, that is, state optimization,
continues to be a bottleneck.  Bundle Adjustment (BA) has been
recognized as the favored back-end solution; it estimates both camera
poses and maps with high accuracy and robustness.  However, BA is
computationally expensive: it optimizes a large number of states with
up to cubic complexity, and the computation process is largely iterative.  

As with the front-end, recognizing the need to improve BA compute time
has led to parallel implementations of BA solving on 
multi-core CPUs \cite{wu2011multicore} and 
GPUs \cite{choudhary2010practical,hansch2016modern}.  These parallel 
BA methods address the time-sensitive nature of the BA problem
(especially for large-scale problems), but they don't fully reduce the
compute and power requirement in BA solving.  
For both compute and power limited configurations frequently seen in
robotics and AR platforms, specialized hardware such as FPGAs may serve
as co-compute components to speed up specific steps in the BA
computations (e.g., FPGA-based Schur elimination \cite{qin2019pi}).  
However the remaining BA steps that don't translate to hardware, such as
re-linearization and factorization, limit the possible compute
reductions of BA-based back-ends.  
Another hardware platform is the graph processor \cite{ortiz2020bundle},
which shows great potential over conventional CPU-based BA solving.
Joint hardware and software acceleration strategies will most likely
provide the best approach to accurate and cost-effective VSLAM back-end
BA solutions.

Algorithm improvements typically seek to replace or reduce the scale of
BA problem in VSLAM system.  
Some state-of-the-art VSLAM systems \cite{li2013high,sun2018robust} choose 
less expensive filters over BA as the back-end solution.  
The computational complexity of filters such as MSCKF \cite{mourikis2007multi} 
is linear in the cardinality of map states.  
However, these filters have the downside of inconsistency and degraded
mapping \cite{WhyFilter}.
The majority of state-of-the-art VSLAM systems utilize BA-based back-ends to achieve 
high performance.  To reduce of the cost of BA, state optimization is typically separated 
into two semi-independent parts: a high-rate, scale-limited 
optimization (local BA) and a low-rate full optimization (global BA).  
The cost-efficiency of a BA-based back-end is largely determined by the high-rate 
local BA.  Several strategies exist to identify scale-limited states
for the local BA problem.  Many systems 
\cite{SVO2017,DSO2017,qin2018vins,liu2018ice,usenko2019visual,Rosinol19arxiv-Kimera} 
only keep recent states (e.g. camera frames and map points) that stay
within a sliding-window.  
Other systems
\cite{strasdat2011double,murORB2,mur2017visual,Campos2020ORBSLAM3AA} 
use the covisibility graph to organize historical states and query the
covisible subgraph for local BA construction.  
However, the heuristic strategies described above do not provide 
insight into the conditioning of the local BA problem.  
In the presence of computational limits, the small subset of 
states selected with these heuristic strategies could form an 
ill-conditioned local BA that is either slow to converge or 
degrades solution accuracy. 

In this paper, we propose a novel, rigorous method to determine 
the state subset in local BA, called {\em Good Graph}, with 
optimization performance guarantees.  
Inspired by recent progress in submodular submatrix selection 
\cite{mirzasoleiman2015lazier,sathanur2018scaling}, we describe 
an efficient algorithm for selecting a subset of states to define a
scale-limited local BA problem that maximizes its conditioning relative
to the full BA problem (Section \ref{sec::goodgraph}).
%
%
We also describe SLAM specific modifications attuned to the needs 
of local BA in the back-end (Section \ref{sec::budget}).
In particular, given that local BA size and compute time are related, we
propose to determine the size of the desired {\em Good Graph} for local BA 
by predicting, on-the-fly, the available optimization time budget to
best support pose estimation in the near future.
A small {\em Good Graph} is selected for the local BA optimization when the
time budget is small (e.g. due to fast motion, or
limited computational resources).  
Otherwise, a large {\em Good Graph} is selected because the time budget can
afford it.  

The proposed {\em Good Graph} algorithm is a form of active problem 
selection for the BA-based VSLAM back-end, which is similar in spirit
to the active matching solution for VSLAM front-ends \cite{zhao2020gfm}.
{\em Good Graph} algorithm predicts future needs, dynamically assigns 
an appropriate size budget, and selects a subgraph by maximizing BA 
estimation conditioning. 
As the scene structure and motion profile change, the {\em Good Graph}
solution will adapt the selection process and subsequent local BA
optimization characteristics.
The {\em Good Graph} algorithm is integrated into a state-of-the-art 
BA-based VSLAM system \cite{murORB2} with a cost-efficient VSLAM 
front-end \cite{zhao2020gfm}.  
The final VSLAM system achieves performance superior to state-of-the-art
VSLAMs within multiple evaluation scenarios and computational limits
(Section \ref{sec::exp}).
The VSLAM system \cite{GG_Code} 
and the full evaluation results
\cite{GG_Results}
are released.  



\section{Related Works} \label{sec::liter}
This section first explores existing efforts on bounding the scale of 
the local BA problem in VSLAM.  Then, algorithmic acceleration for BA
solving is reviewed.  Last, literature related to submodular submatrix 
selection, which serves as the theoretical basis behind {\em Good Graph}, is covered.

\subsection{Bounding Local BA in VSLAM}
VSLAM with a BA-based back-end has better accuracy and robustness than 
VSLAM with a filter-based back-end, as determined by the comparative
study \cite{WhyFilter}.  
Using BA in VSLAM, especially for the high-rate local optimization
module, requires bounding the scale of states to be optimized to achieve
real-time or near real-time optimization rates. The bounding leads to a
local BA problem versus performing a full BA on all measurements to date.  

One prevalent means to limit the problem size is to employ a sliding
window strategy for the local BA 
\cite{SVO2017,DSO2017,qin2018vins,liu2018ice,usenko2019visual,Rosinol19arxiv-Kimera}.
Only recent states (camera frames and map points) within the sliding
window are optimized in local BA.  
Older states outside the sliding window are either 
dropped \cite{leutenegger2015keyframe,SVO2017} or 
fixed as linear priors 
\cite{DSO2017,qin2018vins,liu2018ice,usenko2019visual,Rosinol19arxiv-Kimera}.  
The sliding window strategy as commonly applied in visual-inertial 
odometry is effective under an infinite tunnel assumption for the world
structure and motion profile; it is not an optimal solution for a VSLAM
trajectory with frequently revisited camera poses within the environment.  
The ability to reuse historical information beyond the sliding window is
limited; for instance, when solely relying on loop closing.  
Fixing historical information to be linear priors introduces bias into
the optimization, therefore leading to inferior performance under
frequent revisits.

A second common strategy for bounding the scale of the optimization states 
is covisibility information.  
Covisibility approximates the amount of mutual information between keyframes 
\cite{strasdat2011double}. 
Ideally, subsets of keyframes with strong mutual covisibility form
well-conditioned, local BA optimization problems.
For rapid queries and updates, state-of-the-art VSLAM systems 
\cite{leutenegger2015keyframe,murORB2,mur2017visual,zhao2019maphash,Campos2020ORBSLAM3AA} 
typically store covisibility information as a graph of historical 
keyframes, that is, as a covisibility graph.  
When compared with sliding-window methods, covisibility graphs encode 
more historical information, which is preferred in general SLAM scenarios.  
In the presence of revisits, a covisibility graph enables querying of
earlier keyframes (and map points) for local BA; in the absence of 
revisits, a covisibility graph acts similarly to a sliding window.
However, covisibility information offers only a rough 
approximation of frame-to-frame mutual information.  
Therefore, the actual conditioning of the optimization problems formed 
from covisibility graphs is not guaranteed.  
In practice, VSLAM systems using covisibility graphs typically
over-select states in local BA, which improves the conditioning but
limits the cost-efficiency of BA-based back-ends.

Yet another strategy to select optimization states involves the
re-projection error \cite{sibley2010vast,chou2019tunable}.  Before each
local BA run, map points with large re-projection error are kept as
optimizable states, while points with small error are treated as
fixed priors.  Although treating small-error points as priors does
reduce the scale of local BA significantly, it may introduce bias to the
size-reduced BA problem.  Furthermore, the cut-off threshold of
re-projection error has a strong impact on subgraph selection, and
remains an open problem on its own.

\subsection{Algorithmic Acceleration of BA Solving}
Apart from bounding the scale of local BA, algorithmic accelerations 
to speed up BA solving have been extensively studied.  
Solving the least squares objective of BA problem typically involves
multiple iterations, with each iteration involving a linearized system.  
Speeding up the linearized solution time would accelerate the BA solving.  
Following this observation led to a method for grouping and collapsing
densely visible factors ({\em fragment}) into single factor, thereby
reducing the matrix-vector multiplications required in each iteration  
\cite{carlone2014mining}.  The concept of {\em fragment} extraction was
subsequently translated to online, local BA \cite{litman2019accelerated}.
Fragments of local BA problem are chosen online with an efficient and
scalable algorithm, then used for the local BA solution.
Points that are less visible are discarded from local BA.  Similar to
the covisibility heuristic, the visibility information is a rough
approximation to the actual information.  The fragments extracted don't
necessarily provide guarantees on the conditioning of the size-reduced local
BA problem.

Another key aspect of accelerating BA solving is exploiting the 
incremental structure of SLAM problem.  
Incremental algorithms can expedite matrix factorization, which is a
computationally involved module in local BA.  
For example, iSAM \cite{ISAM} uses Givens rotations to update the QR
factorization incrementally; iSAM2 \cite{iSAM2} further eliminates 
the periodic batch steps for variable re-ordering and relinearization 
with incremental alternatives, guided by Bayes tree representations.  
Other work introduced incremental updates
to the Cholesky factorization in BA-based back-ends \cite{polok2013incremental}.
Incremental algorithms for another computationally involved calculation,
Schur elimination, have also been studied \cite{ila2017fast}.  
The combination of sliding-window and incremental algorithms improves
local BA runtime \cite{liu2018ice}.  
The method presented in this paper has dependency upon the incremental
Cholesky factorization work \cite{polok2013incremental}.  
However, the goal of our work is complementary to these incremental BA
algorithms.  It pursues efficient algorithms to formulate scale-limited
BA problems, while incremental approaches aim to solve a sequence of
related BA problems efficiently.  The cost-efficiency of the local BA
problem will be further improved by combining the proposed BA
formulation and incremental solving.  

\subsection{Submodular Submatrix Selection}
A key aspect of this work is connecting state subset (or subgraph) selection 
to submodular submatrix selection.  Generally, submatrix selection is
NP-hard.  However, shortcuts exist when the objective of submatrix
selection is submodular.  The submodularity of various spectral
preservation objectives has been proven with regards to 
submatrix-selection \cite{shamaiah2010greedy,jawaid2015submodularity}.  
Submatrix selection with a submodular (and monotone increasing) objective 
can be approximated by a greedy method with an approximation guarantee.

Prior work exists regarding speeding up SLAM modules by selecting a
subset of states based on a submodular objective function.  
Initially, submodularity and selection led to accuracy improving and
nearly cost-neutral
implementations \cite{zhang2015good,zhao2018good2}, or to accuracy
improving but more costly implementations \cite{carlone2019attention} 
(but still faster than combinatorial selection). 
The important idea advanced was the value of submodular objectives to
facilitate data-association between keypoints detected in the front-end
and map points from the back-end. The idea then led to the design
of a cost-saving active feature matching algorithm based on maximizing a
submodular objective with accuracy neutral outcomes \cite{zhao2020gfm}.  
The active matching front-end was demonstrated to the reduce computational 
load while preserving pose estimation performance.  
Submodularity to reduce the cost or size of the general BA problem 
has also been explored.  
The 2D pose graph solution and its optimality was related to graph
connectivity, which was quantified by the weighted number of spanning
trees (WST) within the graph \cite{khosoussi2019reliable}.
By proving that WST is monotone log-submodular, a greedy solution was 
presented to find a subgraph that approximately optimizes the WST.  
Our work is similar to \cite{khosoussi2019reliable} for subgraph
selection in BA problems using submodular objective functions.  
However, the type of BA problem targeted in our work is the
time-sensitive local BA of 3D SLAM, while the problem considered in
\cite{khosoussi2019reliable} is primarily the global optimization of 
a 2D pose graph.  
Accordingly, the subgraph selection algorithm proposed in this paper
deals with a more densely connected graph structure, and has a stricter
computate budget for selection.  
To that end, we work directly with the conditioning of the optimization
problem rather than with the graph analog, which provides improved
cost-efficiency of the local BA problem in a state-of-the-art VSLAM
system when applied to practical and challenging scenarios.

To further speed up the greedy selection process under submodular objective
functions, randomized acceleration may be incorporated into the
greedy submatrix selection iterations.  
The greedy method with randomized acceleration, which we call {\em
lazier greedy}, has near-optimal performance guarantees 
\cite{mirzasoleiman2015lazier}.  
There are incremental and distributed implementations of the lazier
greedy method \cite{sathanur2018scaling}.  
Introducing the lazier greedy algorithm into the 
VSLAM front-end through an approach called {\em Good Features} matching, 
provided sufficient time savings to run the selection process in real-time
\cite{zhao2020gfm}.  Compared to the local BA problem of the back-end,
the data-association submatrix selection problem of {\em Good Features}
matching is relatively small.
The matrix processed in this work is the system matrix of a local BA
problem, which easily reaches thousands of rows and columns.   The
proposed {\em Good Graph} algorithm extends the lazier greedy idea with
key improvements tailored to the local BA of VSLAM back-ends.

\section{Preliminary} \label{sec::prelim}
Consider the least squares objective of a general BA problem: 
\begin{equation}  \label{eq:LeastSquare}
  \argmin_{\mathbf{x_c}, \mathbf{x_p}} \sum_{i,j}{\left\Vert \rho(\mathbf{x_c}(i), \mathbf{x_p}(j)) \right\Vert^2_{\mathbf{\Sigma_{ij}}}}, 
\end{equation}
where $\mathbf{x_c}$ is the vector of camera states, 
$\mathbf{x_p}$ is the vector of map states, $\rho$ is 
the residual function (e.g. the on-frame distance between 
measurements and world-to-frame projections).  The covariance 
of each residual term, $\mathbf{\Sigma_{ij}}$, is included when
available.  

Numerical solutions to the non-linear objective \eqref{eq:LeastSquare} 
commonly employ iterative methods based on the Gauss-Newton or
Levenberg-Marquardt algorithms.  Each iteration computes a linear
approximation of the original objective within the trust-region around
the current estimate:
\begin{equation}  \label{eq:LinearSystem}
  \argmin_{\boldsymbol{\delta}} \left\Vert \mathbf{J} \boldsymbol{\delta}-\mathbf{b} \right\Vert^2.
\end{equation}
Solving the linear system \eqref{eq:LinearSystem} is equivalent to
solving the normal equation
\begin{equation}  \label{eq:NormalEq}
  \boldsymbol{\Lambda} \boldsymbol{\delta} = \boldsymbol{\eta}, 
\end{equation}
where $\boldsymbol{\Lambda} = \mathbf{J^T} \mathbf{J}$ and $\boldsymbol{\eta} = \mathbf{J^T} \mathbf{b}$.
The spectral property of the system matrix $\boldsymbol{\Lambda}$ is 
important for two reasons: 
1) a well-conditioned $\boldsymbol{\Lambda}$ suggests fast convergence
for iterative solvers and 
2) the volume of $\boldsymbol{\Lambda}$ is connected to the
information or uncertainty level of the corresponding BA problem.  

For ease of specification and interpretation, the BA problem is
sometimes represented as a factor graph \cite{dellaert2017factor}.  
To support interpetation of the {\em Good Graph} approach, a factor graph
interpretation will be used, with the following terminology:
{\em vertices} represent state entities (e.g. a camera or map point), 
{\em edges} represent measurements, and 
the term {\em graph}, representing the BA problem, consists of the {\em vertex} 
and {\em edge} sets.  

Due to the (worst case) cubic complexity of BA solving, working on a 
subproblem of the original BA with a smaller scale could be more 
cost-effective if the full BA solution is not required.  
We are therefore interested in selecting a subgraph of the full graph
(i.e. of the full BA problem).  As discussed earlier, the spectral
properties of the system matrix are important for BA solving.  It is
desirable to select a subgraph with less states while preserving the
spectral properties of the corresponding system matrix.  


One important spectral property of a system matrix 
is quantified by {\em logDet}, which has been applied 
to guide the feature selection \cite{carlone2019attention} 
and active matching \cite{zhao2020gfm} of VSLAM.  
Compared to other metrics such as the {\em condition} and 
{\em minimum eigenvalue}, {\em logDet} aligns with the goal of 
cost-efficiency.  
It is strictly submodular, which enables efficient selection algorithms.  
Computing {\em logDet} involves Cholesky factorization, which is cheaper
than other commonly involved factorizations such as QR and SVD.
Finally, {\em logDet} captures the necessary spectral properties of
matrix conditioning for solving linear systems of equations.

Using the {\em logDet} metric, the {\em Good Graph selection} objective is
formulated as a submatrix selection problem:
\begin{equation} \label{eq:Combine_Opt}
  \max_{\mathbf{S}\subseteq\{0,1,...,m+n-1\}, |\mathbf{S}|=k} \log \det([\boldsymbol{\Lambda}(\mathbf{S})]),
\end{equation}
where the complete system matrix $\boldsymbol{\Lambda}$ contains 
$m$ camera states and $n$ map states, $\mathbf{S}$ is the index 
subset of the selected camera and map states, 
$[\boldsymbol{\Lambda}(\mathbf{S})]$ is the corresponding submatrix, 
and $k$ is the cardinality constraint.  
The choice of states (vertices) is optimized with the submatrix
selection objective \eqref{eq:Combine_Opt}, while the choice of non-zero
fillings (edges) is conducted implicitly.  The selection process finds a
subgraph with fewer vertices than the full graph, but which retains the
connectivity of the chosen vertices; the sparse structure of the
subgraph is preserved.

\section{Good Graph Selection in General BA} \label{sec::goodgraph}

\begin{figure*}[!htb]
  \centering
  \subfloat[Selection on complete system (graph \& matrix view) \label{subfig-1:dummy}]{%
  	\includegraphics[clip, trim=0cm 2cm 4cm 0cm, width=0.54\textwidth]{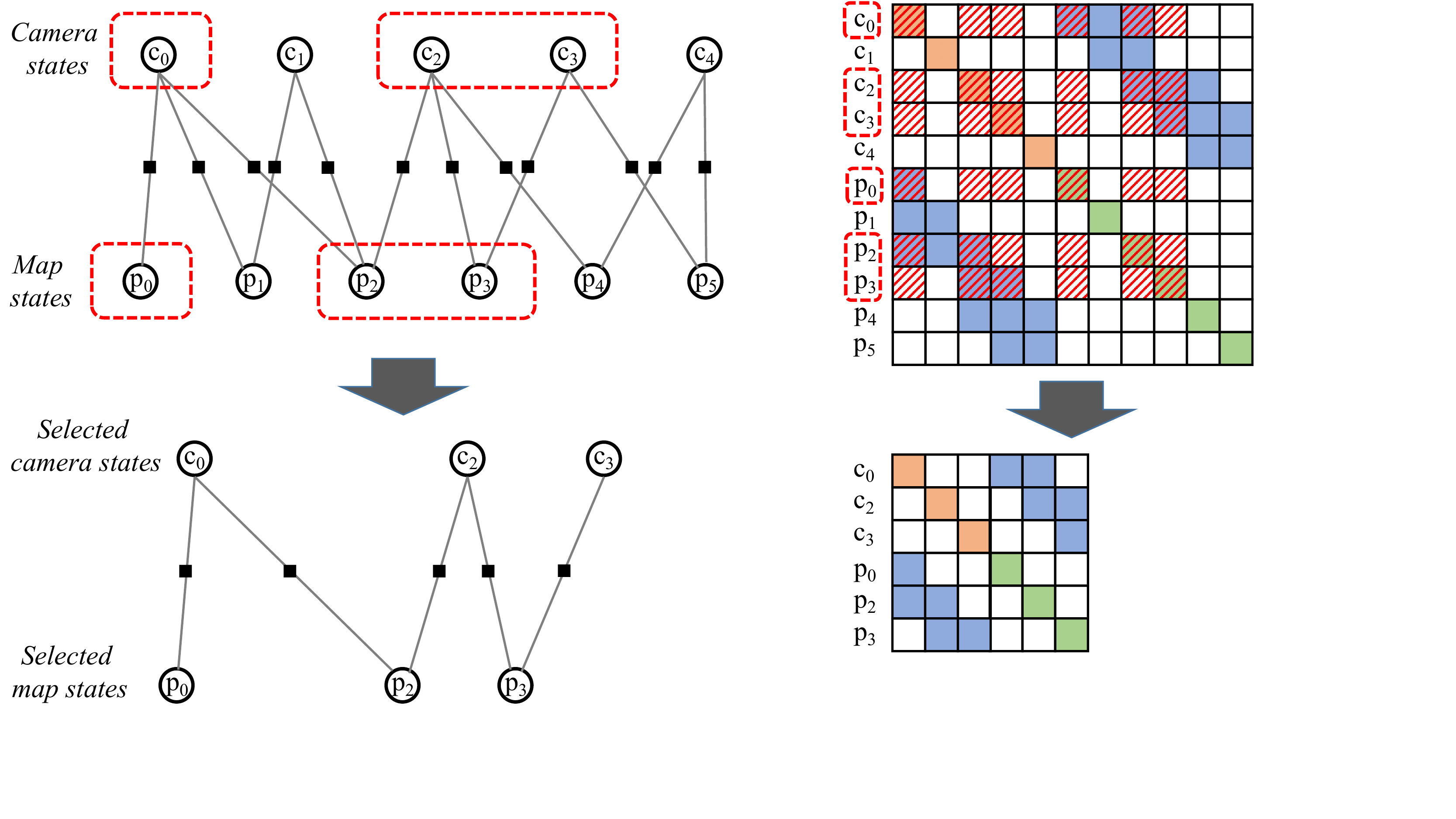}
  }
  \hfill
  \subfloat[Selection on camera-only system (graph \& matrix view)\label{subfig-2:dummy}]{%
  	\includegraphics[clip, trim=0cm 2cm 8cm 0cm, width=0.44\textwidth]{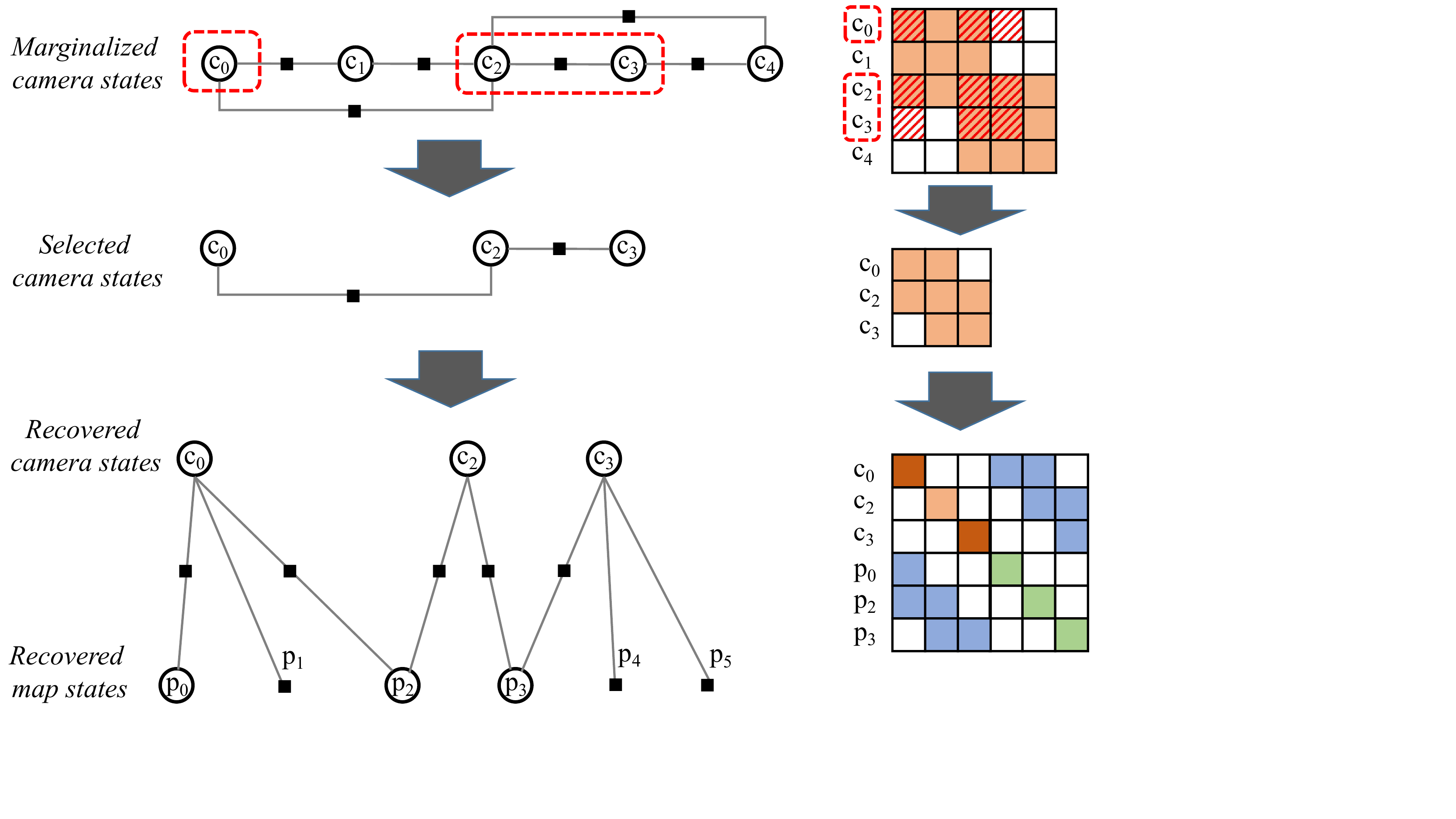}
  }
  \caption{Toy example of subgraph selection on a complete system vs. 
  camera-only system. 
  Working with a camera-only system is desirable for improved efficiency.  
  Compared with the ideal subgraph selected from the complete system, 
  the subgraph selected from the camera-only system (and recovered to include map states) 
  will include priors of the points measured once by the selected cameras, 
  e.g., points $p_1$, $p_4$ and $p_5$.
  In general BA, we simply discard all priors from the selected subgraph.  
  In BA-based VSLAM, we keep the priors of points optimized at least
  once in previous BA runs.
  \label{fig:GoodGraph_Illu}} 
\end{figure*}

The optimal submatrix selection objective \eqref{eq:Combine_Opt}
finds an optimal subgraph (i.e., {\em Good Graph}).  
Ideally, the result will lead to a well-conditioned submatrix, as 
illustrated in the second column of Fig. \ref{fig:GoodGraph_Illu}.  
The corresponding subgraph, visualized in the first column of Fig.
\ref{fig:GoodGraph_Illu}, meets the cardinality constraint and maximizes
{\em logDet}.  However, the problem described by \eqref{eq:Combine_Opt}
is an NP-hard combinational optimization problem, whose solution
will usually be more expensive to solve than the original least-squares
BA.  Instead of tackling the original problem \eqref{eq:Combine_Opt}, 
two relaxations are proposed to reduce the computational cost while
providing sub-optimality guarantees, plus the incremental structure of
the problem is exploited to avoided repeated computation. 

\subsection{Subgraph Selection on Camera-only System}

In practice, there are issues working with the complete system matrix
$\boldsymbol{\Lambda}$.  The large size of $\boldsymbol{\Lambda}$ 
negatively impacts submatrix selection efficiency.  
Additionally, submatrix selection from the joint set of 
camera and map point states may create undesirable behavior due to the relationship
between map points and camera views \cite{carlone2019attention}.  
A common practice in bundle adjustment is to marginalize out the map
states with Schur elimination.  The marginalized matrix will involve
only camera states: 
\begin{equation} \label{eq:Schur}
  \mathbf{M} = \boldsymbol{\Lambda_{cc}} 
      - \boldsymbol{\Lambda_{cp}} \boldsymbol{\Lambda_{pp}^{-1}} 
        \boldsymbol{\Lambda_{cp}}^T, \quad 		
  \boldsymbol{\Lambda} = \left[ \begin{matrix}
	\boldsymbol{\Lambda_{cc}}   & \boldsymbol{\Lambda_{cp}} \\
	\boldsymbol{\Lambda_{cp}}^T & \boldsymbol{\Lambda_{pp}} \\
	\end{matrix} \right].
\end{equation}
The top of the fourth colum of Fig. \ref{fig:GoodGraph_Illu} shows an
example of a marginalized matrix $\mathbf{M}$ with the corresponding
camera-only graph at the top of the third column.  The size of $\mathbf{M}$ 
is smaller than that of $\boldsymbol{\Lambda}$.
The original submatrix selection objective \eqref{eq:Combine_Opt} is
replaced by the submatrix selection objective of the marginalized,
camera-only matrix $\mathbf{M}$: 
\begin{equation} \label{eq:Camera_Opt}
  \max_{\mathbf{S}\subseteq\{0,1,...,m-1\}, |\mathbf{S}|=k} 
    \log \det([\mathbf{M}(\mathbf{S})]),
\end{equation}
where $k$ constrains the camera states only.  
Objective \eqref{eq:Camera_Opt} for a camera-only system is not
equivalent to the original objective \eqref{eq:Combine_Opt}.  
Ideally, a subgraph selected with \eqref{eq:Combine_Opt} would have 
better conditioning, since both camera and map states could be selected 
explicitly.  
However, optimizing \eqref{eq:Combine_Opt} is expensive, operates over
different states with different properties, and may lead to
inconsistency during the optimization.  Map states are selected
implicitly in the more efficient and consistent camera-only objective
\eqref{eq:Camera_Opt}.  All map points visible to the selected camera
subset are taken.  The camera state selection process is visualized in
the second row of Fig. \ref{fig:GoodGraph_Illu}(b).  The third row shows
the recovered camera and map subgraph obtained by including all map
points visible to the selected cameras. 

At this juncture, there is a choice to make regarding the map points
visible to only a single camera from the subset: keep them or remove
them.  When kept, the map points measured by only a single selected
camera view serve as priors.  Their effect will depend on the structure
of the BA problem created by the selected subgraph.
General large-scale BA solves an over-determined system, where the loss
of the priors will not significantly impact {\em logDet} maximization.
These priors may introduce bias as any measurement error associated to
the measurements cannot be corrected.  To avoid introducing bias to the
subgraph (and downstream BA), these map points are ignored and not used
to build the BA subgraph.
As will be shown later in Fig.~\ref{fig:SfM_simulation}(c), 
the {\em logDet} value drop for the final subgraph obtained by
discarding single-view priors is relatively small.
On the other hand BA-based VSLAM back-ends solve a smaller sized BA
problem. Removing all map point priors will have a relatively larger
impact on the solution.  Furthermore, the bias introduced by map point
priors will not be as detrimental since the BA-based back-end improves
map state estimation periodically. Keeping track of whether a map
point has been optimized or not by local BA iterations permits selective
use of map points as priors.
In VSLAM back-end implementations of {\em Good Graph}, priors from map
points that have been optimized at least once before will remain in the
final subgraph. Map point priors that have not yet been processed by a
local BA cycle are ignored.

\subsection{Submatrix Selection with Lazier Greedy}
Though the camera-only objective \eqref{eq:Camera_Opt} involves
combinatorial selection from a smaller set, it is still NP-hard.
To efficiently solve the camera-only objective while limiting the loss
in optimality, the submodularity of the {\em logDet} set function is
exploited. The {\em logDet} function is submodular and monotone
increasing \cite{shamaiah2010greedy}. 
Submodularity is extremely useful. It indicates that the combinatorial
optimization objective \eqref{eq:Camera_Opt} can be approximately solved
with greedy methods. 

\begin{theorem} \cite{nemhauser1978analysis} \label{submodGreedy}
\label{submodular}
Given a normalized, monotone, submodular set function $\mathbf{f}: 2^F
\rightarrow \mathbb{R}$, whose optimal solution to the maximization
problem \eqref{eq:Camera_Opt} is denoted by $S^*$\!\!,\; then the set
$S^\#$\!\!\!,\;
computed by the greedy heuristic, is such that:
\begin{equation} \nonumber
  \mathbf{f}(S^\#) \geq (1 - 1/e)\mathbf{f}(S^*) \approx 0.63\mathbf{f}(S^*).
\end{equation}
\end{theorem} 

The bound in Theorem \ref{submodGreedy} ensures that the worst-case
performance of a greedy algorithm cannot be far from optimal.
The $(1-1/e)$ approximation ratio is the best ratio any polynomial
algorithm can achieve (assuming $P \neq NP$) \cite{feige1998threshold}.

The camera-only objective \eqref{eq:Camera_Opt} can be solved with 
a greedy method as follows.
Starting from the submatrix $\mathbf{M}(0)$ of an initial camera vertex 
(i.e., the most recent keyframe), the algorithm iteratively searches for 
the best submatrix that has one more state than $\mathbf{M}(0)$.  
After $k-1$ iterations, the submatrix contains $k$ camera 
states and selection stops.  Finally, all map states that are 
visible by the chosen cameras are included.  The computational 
complexity of greedy method is $\mathcal{O}(km)$.

Inspired by earlier active matching work \cite{zhao2020gfm}, a lazier 
greedy algorithm \cite{mirzasoleiman2015lazier} is utilized 
to further improve the solve time of \eqref{eq:Camera_Opt}.  
In contrast with the greedy method, the lazier greedy method only evaluates 
a random subset of candidate states (row and column blocks) 
at each iteration.  The size $s$ of the random candidate subset 
is controlled by a decay factor $\epsilon$: $s=\frac{m}{k}\log(\frac{1}{\epsilon})$. 
The computational complexity of the lazier greedy method 
is $\mathcal{O}(\log(\frac{1}{\epsilon})m)$, 
which is lower than the $\mathcal{O}(km)$ complexity of the 
greedy method. 
Furthermore, the approximation ratio of the lazier greedy method 
has an $\epsilon$-controllable drop from that of the greedy method:
\begin{theorem} \cite{mirzasoleiman2015lazier}
\label{optimal_in_exp}
Let $f$ be a non-negative monotone submodular function and set 
$s=\frac{m}{k}\log(\frac{1}{\epsilon})$. 
Then lazier greedy achieves a $\left( 1-1/e-\epsilon \right)$ approximation
guarantee in expectation to the optimum solution of \eqref{eq:Camera_Opt}.
\end{theorem} 
\begin{theorem} \cite{hassidim2017robust}
\label{optimal_in_prob}
The expected approximation guarantee of $\left( 1-1/e-\epsilon \right)$
is reached with a minimum probability of $1-e(-0.5k(\sqrt{\mu}+ln(\epsilon+e^{-1})/\sqrt{\mu})^2)$, 
when maximizing a monotone submodular function under the cardinality constraint $k$ with lazier greedy.  
$\mu \in (0,1]$ is the average approximation ratio when maximizing
the margin gain during each iteration of lazier greedy.
\end{theorem}

The $\epsilon$ drop balances the speed of execution versus the
approximation guarantee. How to mediate these two will depend on the
problem being solved and the typical sizes of $k$ and $m$.
For all experiments performed in later sections, the decay factor
$\epsilon$ is set to $0.0025$ (i.e., $1/400$). With this value,
lazier greedy selection has a logarithmc drop in computational cost, 
while preserving the approximation guarantee of the optimal solution 
(a small linear drop in expectation, with a minimum probability close to
1). The expected runtime is $\mathcal{O}(6.0 m)$ versus
$\mathcal{O}(km)$ for the greedy method, with typical values of $k$
ranging within $20-60$ during local BA optimization by VSLAM back-ends.

\subsection{LogDet with Incremental Cholesky}
One bottleneck of the lazier greedy algorithm is the 
cost of computing the {\em logDet} metric.  For the positive definite 
square matrix $\mathbf{M}$, efficient computation of {\em logDet} 
involves Cholesky factorization. 
For $\mathbf{M} = \mathbf{L} \mathbf{L}^T$, 
$\log \det(\mathbf{M}) = 2 \sum{\log(diag(\mathbf{L}))}$.  
However, simply using a Cholesky-based {\em logDet} computation 
within the lazier greedy algorithm will not provide a speed improvement.
The size of the selected submatrix grows linearly as a function of the 
lazier greedy iterations, while the Cholesky factorization cost grows
cubically, which affects the cost-efficiency of submatrix selection.

Given that the submatrix is incrementally built during the selection
process, the Cholesky factorization in the {\em logDet} computation 
should likewise be obtained incrementally. Doing so reduces the cost
to evaluate {\em logDet}. Therefore, with each iteration of the lazier
greedy algorithm, the system matrix of the current selection, dubbed
$\mathbf{M}(i)$, will be partially updated.  The partial updates involve
appending the updated blocks to the rows and columns of the existing
matrix $\mathbf{M}(i)$:
\begin{equation} \label{eq:Inc_Submatrix}
\mathbf{M}(i+1) = \left[ \begin{matrix}
	\mathbf{M}(i) 	& \mathbf{B} \\
	\mathbf{B}^T 	& \mathbf{D} \\
	\end{matrix} \right].
\end{equation}
Given the Cholesky factorization $\mathbf{L}(i)$ of $\mathbf{M}(i)$, 
the Cholesky factorization of new submatrix $\mathbf{M}(i+1)$ is
\cite{osborne2010bayesian}: 
\begin{equation} \label{eq:Inc_Cholesky}
\begin{aligned} 
\mathbf{L}(i+1) &= \left[ \begin{matrix}
	\mathbf{L}(i) 	& \mathbf{L_{1}} \\
	\mathbf{0} 		& \mathbf{L_{2}} \\
	\end{matrix} \right], \\
\mathbf{L_{1}} &= (\mathbf{L}(i)^{+})^T \mathbf{B}, \\ 
\mathbf{L_{2}} &= chol(\mathbf{D} - \mathbf{L_{1}}^T \mathbf{L_{1}}).
\end{aligned}
\end{equation}
Computing {\em logDet} with the incremental formula \eqref{eq:Inc_Cholesky} 
further improves the cost-efficiency of Good Graph selection.

\subsection{Implementation Details and Validation \label{sec:ImpDet_Valid}}

Besides the three algorithmic improvements just described, the base Good
Graph algorithm includes one additional improvement, while the
SLAM-focused Good Graph algorithm further adds one more, with both based
on good engineering practice. 
The former involves using an analytical form of the Jacobian 
for generating the system matrix $\boldsymbol{\Lambda}$ 
\cite{sola2017quaternion}. Having a closed-form function for the
Jacobian is more efficient than using numerical schemes. The latter
exploits the fact that typical VSLAM back-ends already limit the local BA
problem size based on covisibility constraints. 
Thus, there is a flag to bound the candidate pool of
camera states based on the covisibility graph as the source candidate
set, which is faster to run subselection for than using all camera
vertices as the source set (due to the 
$\mathcal{O}(\log(\frac{1}{\epsilon})m)$ cost). 
The covisibility bounded {\em Good Graph} sets the flag to true.  All of
these elements comprising the {\em Good Graph} algorithm were integrated
into a state-of-the-art BA solver, SLAM++ \cite{ila2017slam++}.  For
less-constrained applications such as general BA, a mixed approach can
be pursued, where the covisibility candidate pool extends to
more camera vertices with second-order or higher neighbors of the
covisibility graph.  

\begin{table}[t]
	\small
	\centering
	\caption{Proposed Improvements Covered by Good Graph Variants}
	\begin{tabular}{|c|ccccc|}
		\toprule
		\bfseries Variants &\em A &\em B &\em C &\em D &\em E \\
		\midrule
Camera-only 		& \cmark  & \cmark  & \cmark  & \cmark & \cmark	\\
Lazier Greedy		& \xmark  & \cmark  & \cmark  & \cmark & \cmark	\\
Incremental Chol. 	& \xmark  & \xmark  & \cmark  & \cmark & \cmark	\\
Analytical Jacob. 	& \xmark  & \xmark  & \xmark  & \cmark & \cmark \\
Covis-bounded Pool	& \xmark  & \xmark  & \xmark  & \xmark & \cmark	\\ 
		\bottomrule	
	\end{tabular} 
	\label{tab:Ablation}
\end{table}

To validate the cost-efficiency of the {\em Good Graph} algorithm, 
three experiments were conducted: two with a general BA dataset, 
the Venice dataset \cite{g2o}, which includes 871 cameras and 530k map
points; 
the other on randomly generated BA problems with 50 cameras and 6k map
points, which simulates the problem of local BA in VSLAM back-end.  
In all three experiments, the BA solver SLAM++ is configured with a 
maximum iteration number of 20.

\subsubsection{General BA Subgraph Selection Ablation Study}
The study with the 871-camera Venice dataset is an ablation type of
study on the {\em Good Graph} algorithm. Five {\em Good Graph} variants,
consisting of incrementally applied modifications to arrive at the full
VSLAM {\em Good Graph} variant are applied to reconstruct the Venice
scene.
The full BA with all camera and map states is also evaluated to serve as
a reference.  Table~\ref{tab:Ablation} describes the incremental
implementation schedule and variant letter assignment.  Each variant
selects an 87-camera subgraph from the full graph, for input to the
SLAM++ BA solver.
All computations are executed on a PC 
(Intel i7-7700K, PassMark 2583 per core).  
Problem size, timing, and error statistics for all 
implementations are reported in Tables \ref{tab:TimeCost_BA} -
\ref{tab:RMSE_BA}.  

Time cost breakdowns of the full BA and the five GG variants are reported in
Table~\ref{tab:TimeCost_BA}.  Initially the time cost increases, which
is the common issue for active approaches that aim to reduce problem size.  
Optimization-based active selection can take longer to run than the time saved. 
However, as additional algorithmic shortcuts
are incorporated, the problem reduction time cost improves.
The time consumption of {\em Good Graph} variant {\em E}, which covers all
improvements described, is the lowest of all six methods.
Each improvement described above had a clear, positive impact on 
the cost-efficiency of {\em Good Graph} selection.  
To understand whether the implementations lead to improved performance,
Table \ref{tab:time} looks at the time cost per map point optimized.
It factors out the effect of the problem size on the solution time cost.
It is not until the last variant (E) that the time cost to solve per point
becomes lower than the full BA time cost per point. 

The map point accuracy of the BA solutions is also evaluated with the
RMSE between the 3D points optimized by the tested implementations and
the ground truth points.  
The fixed number of iterations in the optimization loop provides a means
to evaluate the convergence efficiency of the methods.  
However, each implementation will optimize different sets of 3D points,
which might serve to confound the RMSE statistics. Therefore,
Table~\ref{tab:RMSE_BA} contains the RMSE of all map points optimized
per variant, and on the subset of points common to all methods, which is
the intersection of 3D map points selected by all five variants.  
%
Though selecting the same number of cameras (i.e., 87) as other
variants, variant {\em E} appears to have a higher percentage of
high-quality points that improve the optimization conditioning based on
having the lowest RMSE error.
Selection from the covis-bounded pool ensures strong covisibility
between selected cameras.  Moreover, it is less likely to take
points with single-views, which are considered to provide a biased prior
and are discarded before BA solving.  

Overall, the full {\em Good Graph} exhibits the best time cost per point
processed and the best RMSE error, which indicates that the full {\em
Good Graph} selection process is efficient and achieves the desired goal
of improving the conditioning of the optimization problem. Though less
points are processed when compared to the {\em Full} problem, this factor is
not problematic. Having well conditioned points to match against during
the local map pose optimization process in the front-end is the main
objective of {\em Good Graph}.  If {\em Good Graph} were to be sought
for the full BA optimization, then additional steps would be required to
incrementally incorporate new camera views and map points in a manner
that does not compromise the existing optimized states. Running
additional iterations of {\em Good Graph} with properly weighted costs
might provide an improved solution over the full optimization, without
compromising run-time.

\begin{table}[!tb]
	\footnotesize
	\centering
	\caption{Time Cost Breakdown (ms) of Good Graph Variants
	    \label{tab:TimeCost_BA}}
	\begin{tabular}{|c|c|cccccc|}
		\toprule
		\multicolumn{2}{|c|}{\bfseries Variants} &\em Full &\em A &\em B &\em C &\em D &\bfseries \em E \\
		\midrule
		\parbox[t]{1mm}{\multirow{5}{*}{\rotatebox[origin=c]{90}{\bfseries Subgraph}}}
&  Jacob.   		&  -    & 6,531   & 6,902  & 6,628 	& 5,787 & 456	\\
&  Schur    		&  -    & 3,409   & 3,499  & 3,426  & 3,408 & 221	\\
&  Chol.    		&  -    & 80,153  & 3,990  & 148    & 131  	& 46	\\
&  Misc.    		&  -    & 28,695  & 110    & 272    & 264  	& 80 	\\
&  Total 			&  -    & 118,788 & 14,501 & 10,474 & 9,590 & 803	\\ 
\midrule
\multicolumn{2}{|c|}{\bfseries BA Solving} & 46k & 31,791 & 17,953 & 17,685 & 5,740 & 2,839 \\ 
\midrule
\multicolumn{2}{|c|}{\bfseries Total Time}  & 46k & 150.1k & 32.5k &
28.2k & 15.3k & 3.6k \\ 
		\bottomrule	
	\end{tabular} 
	\vspace*{0.1in}
	\caption{Time Cost (Selection \& BA) per Point Processed ($\mu$s/\#) \label{tab:time}}
	\begin{tabular}{|c|cccccc|}
		\toprule
		{\bfseries Variants} &\em Full &\em A &\em B &\em C &\em D &\bfseries \em E \\
		\midrule
        Time & 86.7 & 1,217.7 & 266.0 & 230.1 & 124.6 & \bf 52.8
        \\ \bottomrule	
	\end{tabular} 
	\vspace*{0.1in}
%
	\small
	\centering
	\caption{Graph Scale and RMSE (mm) of Bundle Adjustment \label{tab:RMSE_BA}}
	\begin{tabular}{|c|c|cccccc|}
		\toprule
		\multicolumn{2}{|c|}{\bfseries Variants} &\em Full &\em A &\em B &\em C &\em D &\bfseries \em E \\
		\midrule
\parbox[s]{1mm}{\multirow{2}{*}{\rotatebox[origin=c]{90}{\bfseries \footnotesize Scale}}}
& Cam. 	& 871   & 87   & 87   & 87  & 87  & 87 \\
& Point & 530k  & 124k & 122k & 122k & 123k & 69k \\
\midrule
\parbox[s]{1mm}{\multirow{2}{*}{\rotatebox[origin=c]{90}{\bfseries \footnotesize RMSE}}}
& All  	& 49.2 & 41.9 & 45.7 & 45.7 & 41.2 &\bf 39.7  \\
& Int. 	& 49.3 & 41.8 & 45.7 & 45.7 & 41.1 &\bf 39.8  \\
		\bottomrule	
	\end{tabular} 

\end{table}

%
%
%

\addtolength{\tabcolsep}{-4pt}    
\begin{figure*}[!t]
  \centering
  \begin{tikzpicture}[inner sep=0pt,outer sep=0pt]
  \node (VD) at (0in,0in)
    {\includegraphics[clip, trim=0cm 12.4cm 1cm 0cm, width=\linewidth]{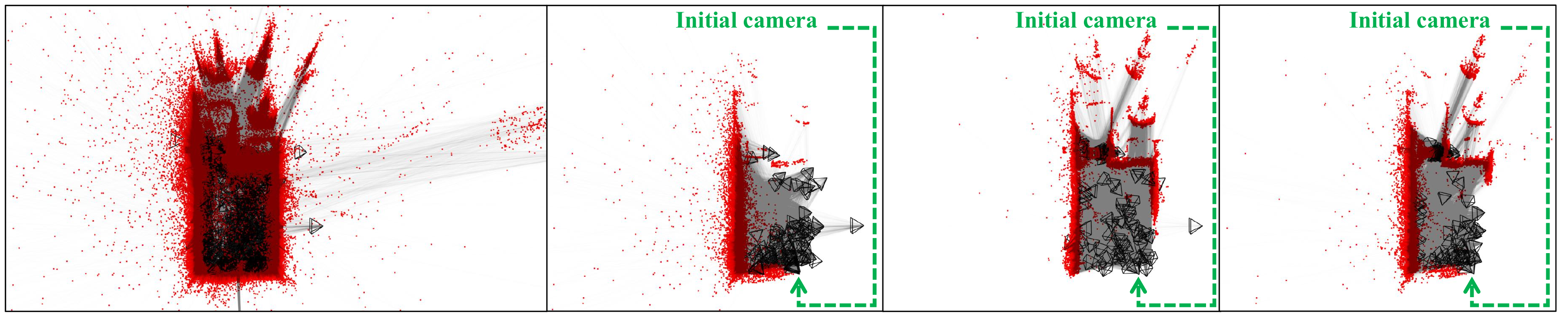}};
    \node[anchor=north west] at ($(VD.south west)+(0.6in,0in)$)
      {\scriptsize \parbox{2in}{\begin{tabular}{p{0.70in}l}
        Full graph: & 871 cameras \\
                    & 530k points \\[2pt]
        Total time cost: & 45.6 sec \\[2pt]
        PCD RMSE (all): & 49.2 mm \\
        PCD RMSE (int): & 47.5 mm \\[2pt]
        Time per point: & 86.7 $\mu$s/pt
        \end{tabular}}};
    \node[anchor=north west] at ($(VD.south west)+(2.5in,0in)$)
      {\scriptsize \parbox{2in}{\begin{tabular}{p{0.810in}l}
        Covisible subgraph: & 87 cameras \\
                    & 47k points \\[2pt]
        Total time cost: & 1.3 sec \\[2pt]
        PCD RMSE (all): & 39.8 mm \\
        PCD RMSE (int): & 39.7 mm \\[2pt]
        Time per point: & 27.7 $\mu$s/pt
        \end{tabular}}};
    \node[anchor=north west] at ($(VD.south west)+(4.05in,0in)$)
      {\scriptsize \parbox{2in}{\begin{tabular}{p{0.810in}l}
        Random subgraph: & 87 cameras \\
                    & 34k points \\[2pt]
        Total time cost: & 2.0 sec \\[2pt]
        PCD RMSE (all): & 46.2 mm \\
        PCD RMSE (int): & 46.6 mm \\[2pt]
        Time per point: & 58.8 $\mu$s/pt
        \end{tabular}}};
    \node[anchor=north west] at ($(VD.south west)+(5.58in,0in)$)
      {\scriptsize \parbox{2in}{\begin{tabular}{p{0.810in}l}
        Good subgraph: & 87 cameras \\
                       & 69k points \\[2pt]
        Total time cost: & 3.6 sec \\[2pt]
        PCD RMSE (all): & 39.7 mm \\
        PCD RMSE (int): & 38.1 mm \\[2pt]
        Time per point: & 52.8 $\mu$s/pt
        \end{tabular}}};
  \end{tikzpicture} 
  \caption{BA example on Venice dataset \cite{g2o}.  
  \textbf{From left to right}: 
  1) BA on full graph;
  2) BA on subgraph selected with covisibility information;
  3) BA on subgraph selected randomly;
  4) BA on subgraph selected with proposed Good Graph algorithm.     
  All three subgraphs start with the same camera (i.e., initial vertex).  
  The time cost of subgraph BA is less than that of full graph BA.    
  Compared with covisible and random subgraphs, BA on the good subgraph
  has better 3D reconstruction accuracy 
  (lower RMSE in the reconstructed point cloud), while 
  processing twice amount of 3D points.  
    \label{fig:RMSE_BA}} 	
\addtolength{\tabcolsep}{4pt}    
	\centering

    \vspace*{0.5em}
    \begin{tikzpicture}[inner sep=0pt,outer sep=0pt]
      \node[anchor=south west] (EvT) at (0in,0in)
	    {\includegraphics[clip, trim=3.6cm 9.65cm 4.1cm 10cm, width=0.36\linewidth]{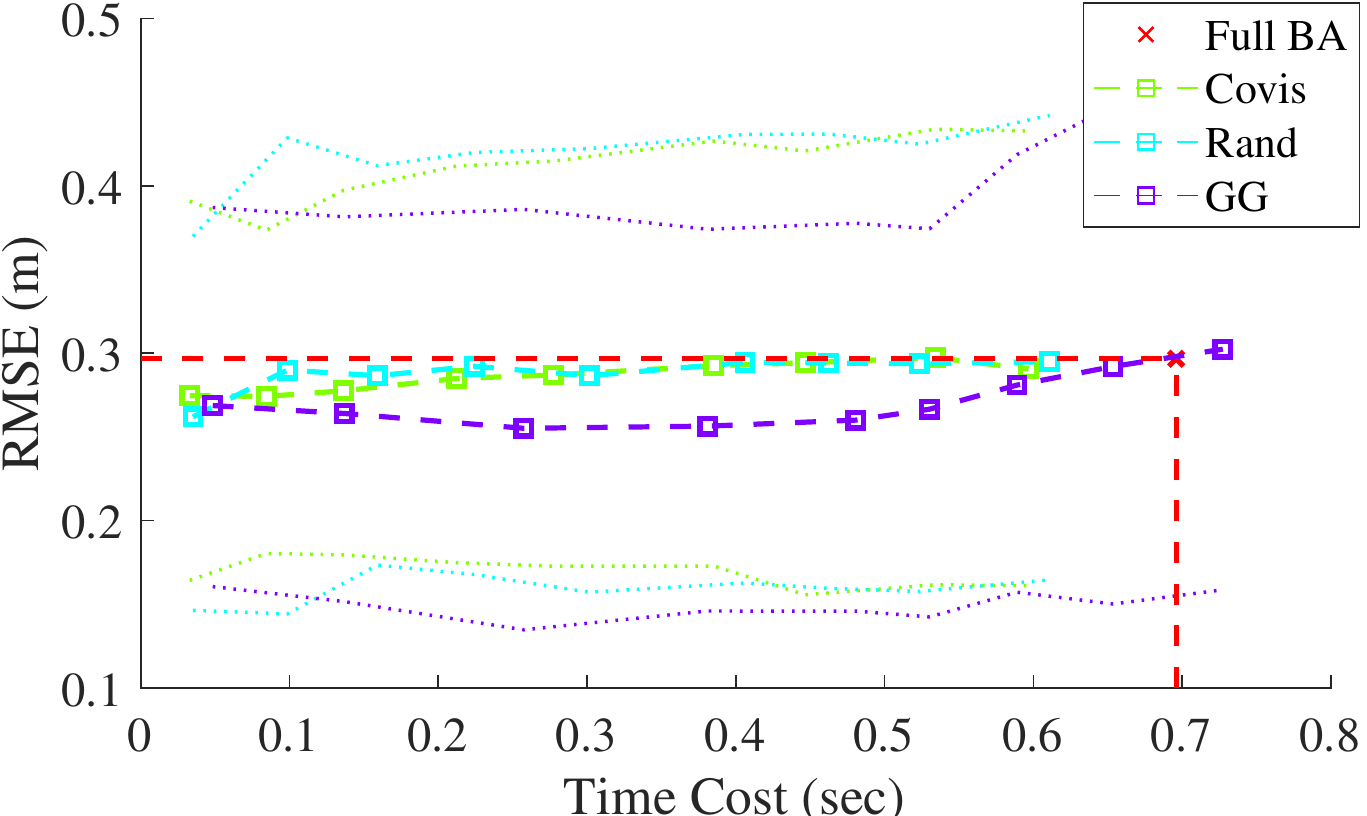}};
      \node[anchor=south west] (PvT) at (EvT.south east)
	    {\includegraphics[clip, trim=3cm 9.75cm 4.2cm 10cm, width=0.38\linewidth]{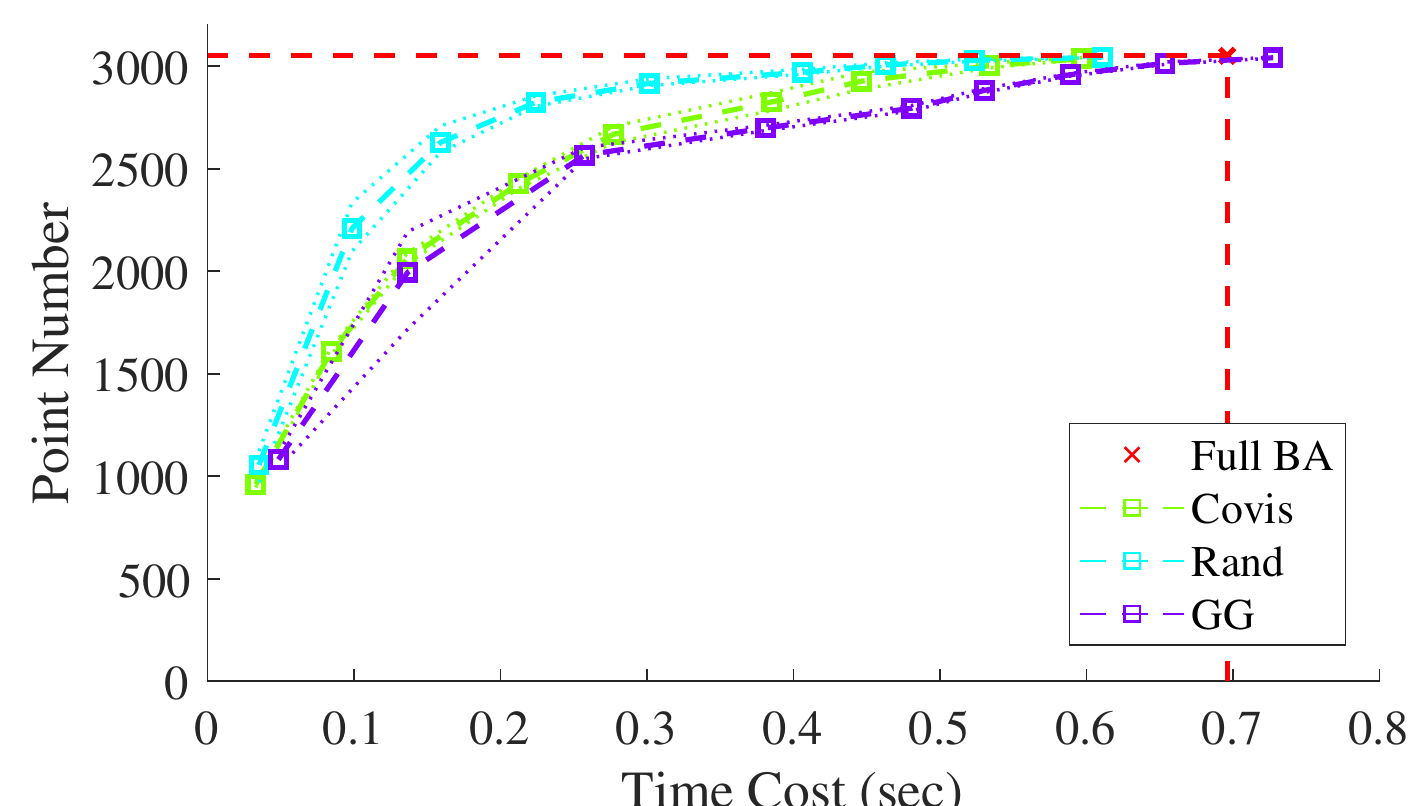}};
      \node[anchor=south west] (LvS) at (PvT.south east)
	    {\includegraphics[clip, trim=5cm 9.6cm 6.5cm 10.2cm, width=0.24\linewidth]{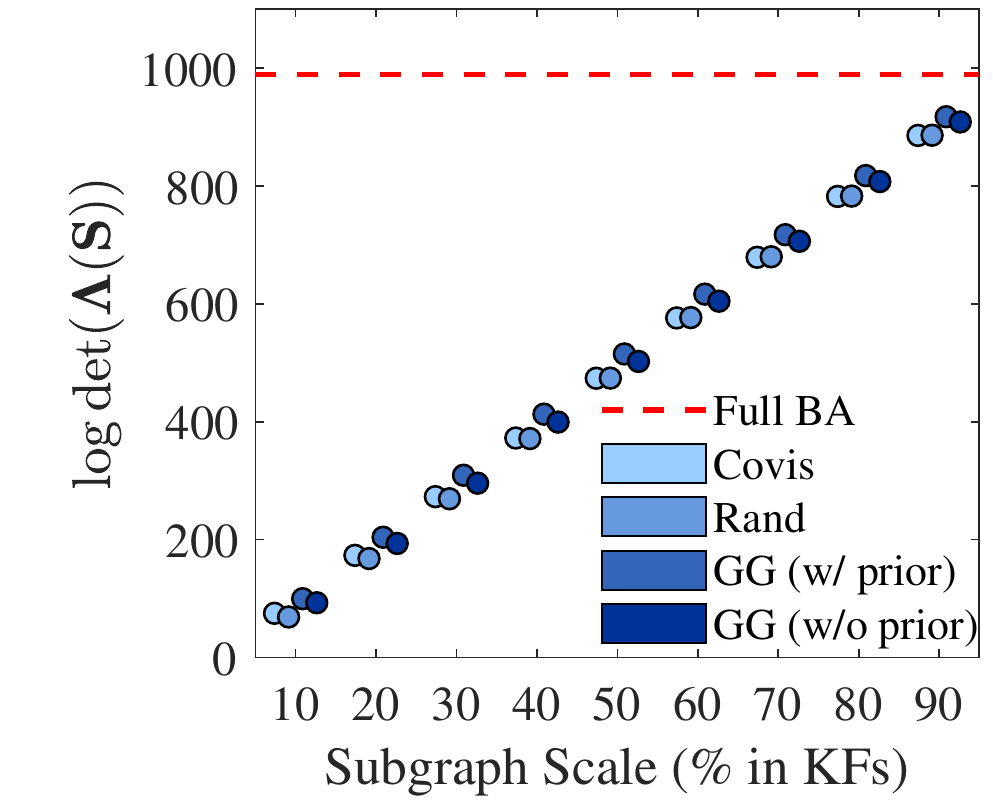}};
    
      \node[anchor=south west] at (EvT.south west) {\small (a)};
      \node[anchor=south west,xshift=7pt] at (PvT.south west) {\small (b)};
      \node[anchor=south west,xshift=7pt] at (LvS.south west) {\small (c)};
    \end{tikzpicture}
  \caption{Cost-efficiency of full and subgraph BA in simulated scenarios.  
  \textbf{Left:} total time cost (of subgraph selection and BA) vs. reconstruction error (RMSE of 3D point cloud) 
  for full BA and three subgraph BA.  
  The average point of full graph BA under 100-repeat is marked with red cross.
  Meanwhile the average curve of subgraph BA under different subgraph scales is in square and dashed line.  
  The first and third quarter curves of subgraph BA in 100-repeat are also plotted with dotted lines.
  \textbf{Middle:} total time cost vs. number of map point optimized in BA, using the same plotting style as previous.
  \textbf{Right:} boxplot of submatrix {\em logDet} scores in subgraph BA, under different subgraph scales.
  The average {\em logDet} of full system matrix is in red dashed line. 
    \label{fig:SfM_simulation}} 	
\end{figure*}

\subsubsection{General BA and Subgraph Selection}
Secondly, the full-featured {\em Good Graph} algorithm (i.e., variant {\em E}) 
is compared against two subgraph selection heuristics: 
1) covisibility-based selection (also called max-covis here), and 
2) random selection. 
Covisibility-based selection first ranks the cameras based on the number
of covisible points shared with the initial camera vertex, then takes
the top-$(k-1)$ cameras and corresponding covisible points to construct
a small BA problem.  Random selection takes camera vertexes randomly
from the full set, as well as the corresponding covisible points
observed by the selected camera subset.  Both are simpler, more 
light-weight selection heuristics than {\em Good Graph} (without
co-visibility bounding).
The outcomes are illustrated in Fig.~\ref{fig:RMSE_BA}, with the RMSE
computed for all points selected by the methods (all), and for the
points all methods have in common (int).
Compared with the two heuristics, {\em Good Graph} selects more points while
achieving lower RMSE after subgraph BA.  As a mixture of spectral
property maximization and covisibility bounding, {\em Good Graph} selection
finds a better subgraph, quantitatively and qualitatively, than the purely
covisibility-driven selection.  Random sampling, on the other hand, can 
be seen as an extreme case of {\em Good Graph} selection: in each iteration of
lazier greedy, only 1 sample is randomly taken for evaluation.  
According to Theorem \ref{optimal_in_exp}, random sampling is 
equivalent to lazier greedy selection with $\epsilon=e^{-\frac{k}{m}}$.  
The approximation ratio of random sampling is $1-1/e-e^{-\frac{k}{m}}$ 
in expectation.  Due to the looser guarantee on performance, random
sampling may end up with the worst performance of the three subgraph
BA methods tested: it may take more time to process per point, and may
have the highest RMSE. Here that does happen.

The plotted point clouds based on the selected cameras in 
Fig.~\ref{fig:RMSE_BA} shows that the {\em Good Graph} selection
picks different points when compared to the covisibility method.
The lower RMSE values for (all) and (int) of the {\em Good Graph}
solution indicates that the selected camera and point combinations have
better conditioning, since the number of optimization
iterations is constant across the implementations. 
The max-logDet objective function can provide better camera selections than
the max-covis and random selection heuristics.  
Meanwhile the time cost per point of {\em Good Graph} lies near the average of the
values obtained for the {\em Full} and {\em Covisibility} implementations
(which is 57.2 $\mu$s/pt). There is the added cost of identifying the
{\em Good Graph} in order to improve the conditioning of the BA problem.
These two competing factors lead to an error vs compute trade-off in
identifying a smaller BA problem. Balancing this within a SLAM system
will require additional modifications to the baseline {\em Good Graph}
implementation.

\subsubsection{Local BA with Sequential Camera Poses}
The last experiment further validates the reconstruction accuracy and BA
conditioning of the {\em Good Graph} algorithm with a randomized
simulation of a small-scale BA problem, designed to resemble the local
BA of VSLAM back-ends.  The simulation scenario includes a 6DoF
trajectory consisting of 50 cameras and a total of 6000 map points.  
The cameras move in a uniform circular motion, while the map points are 
randomly instantiated in the scenario.  To guarantee the conditioning 
of the constructed BA problem, the following conditions are enforced: 
each point is visible from at least 2 cameras, and 
each camera observes at least 20 points.
Subgraphs are selected with 9 desired cardinalities: from 10\% to 90\%
of cameras from the 50 cameras. 
Under each desired cardinality, a subgraph is selected using the 
proposed {\em Good Graph} algorithm ({\em GG}) without covisibility
bounding, and solved as a small BA problem.  
Two heuristics, as introduced in the previous experiment, are included as well: 
covisibility-based selection ({\em Covis}), 
and random selection ({\em Rand}).  


The simulation results of 100 random configurations are summarized in 
Fig.~\ref{fig:SfM_simulation}.  At the extremes (e.g., very few points or
almost all points), the methods have similar performance, c.f.,
Fig.~\ref{fig:SfM_simulation}(a). The $logDet$
versus graph scale plot qualitatively shows less separation of
conditioning between the three methods, while there is a more easily
seen gap between {\em Good Graph} conditioning and that of the other
methods for subgraph scales between 30-70\%, which manifests as lower
RMSE for those scale values. For most cases local BA with {\em GG} has
better accuracy than the other two heuristics, given the same amount of
time for subgraph selection and BA solving.  
In essence, there exist {\em GG} configurations that can outperform
random and covisibility selection for non-extremal selection
percentages.  When the desired subgraph cardinality reaches 80\% or
higher, the RMSE of {\em GG} BA increases to the same level of the other
subgraph BA methods and the full BA.  In such cases there is little 
value in using {\em GG}.
While this observation suggests some limits to {\em GG} improvements for 
high subgraph cardinality, it is less of a concern in practice; most
{\em GG} calls in SLAM will be triggered with subgraph cardinalities
below 80\%.

The analysis provides further indication that the camera states chosen
by {\em GG} generate a fundamentally different set of map points than
those from {\em Covis}, with chosen camera states that may have poor
covisibility but strong baseline matches. 
The {\em GG} approach is also not trying to maximize map points since
random selection leads to larger BA
configurations (except at 10\%), per Fig.~\ref{fig:SfM_simulation}(b). 
Additional support for the assertion lies in the {\em logDet} scores for
the selected submatrices using different subgraph selection methods,
c.f.  Fig.~\ref{fig:SfM_simulation}(c).
Under each desired subgraph cardinality, the submatrix selected with
{\em GG} has higher {\em logDet}, which indicates the effectiveness of
the {\em GG} algorithm at finding camera poses with good overall BA
conditioning.  The improvement holds regardless of whether the priors
for map points with single camera measurements are included or ignored,
with inclusion giving better conditioning.


\section{Budget-Awareness of Local BA in VSLAM} \label{sec::budget}

\begin{figure*}[!htb]
  \centering
  \includegraphics[clip, trim=0cm 12.2cm 11.5cm 0cm, width=0.495\linewidth]{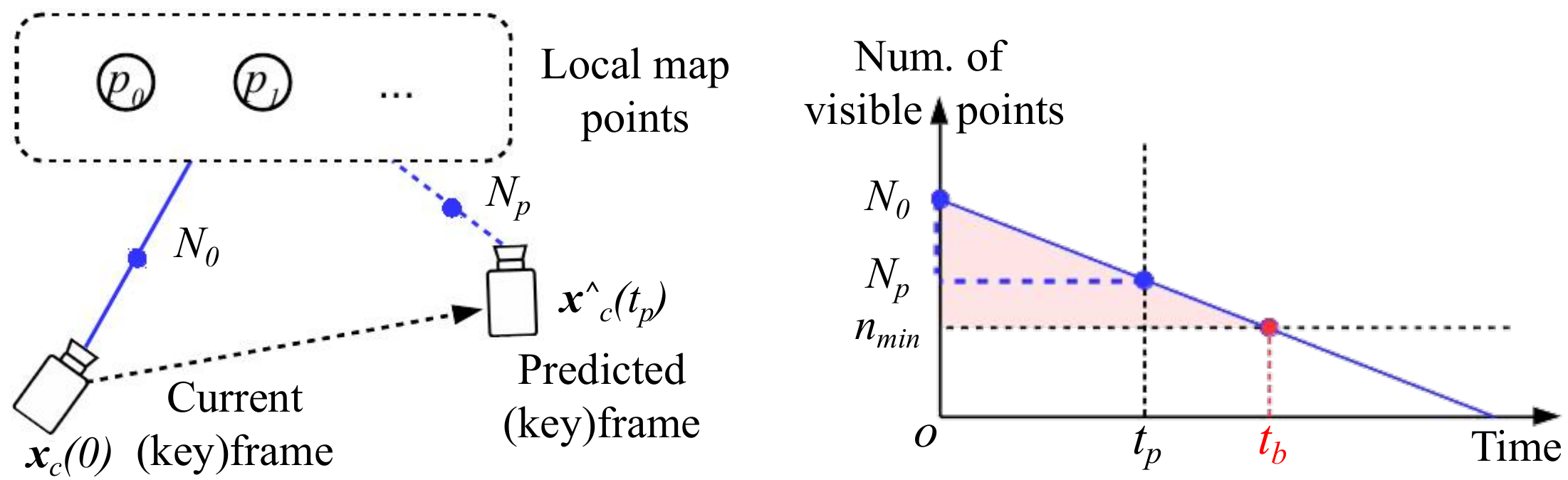} 
  \includegraphics[clip, trim=0cm 12.3cm 11.5cm 0cm, width=0.495\linewidth]{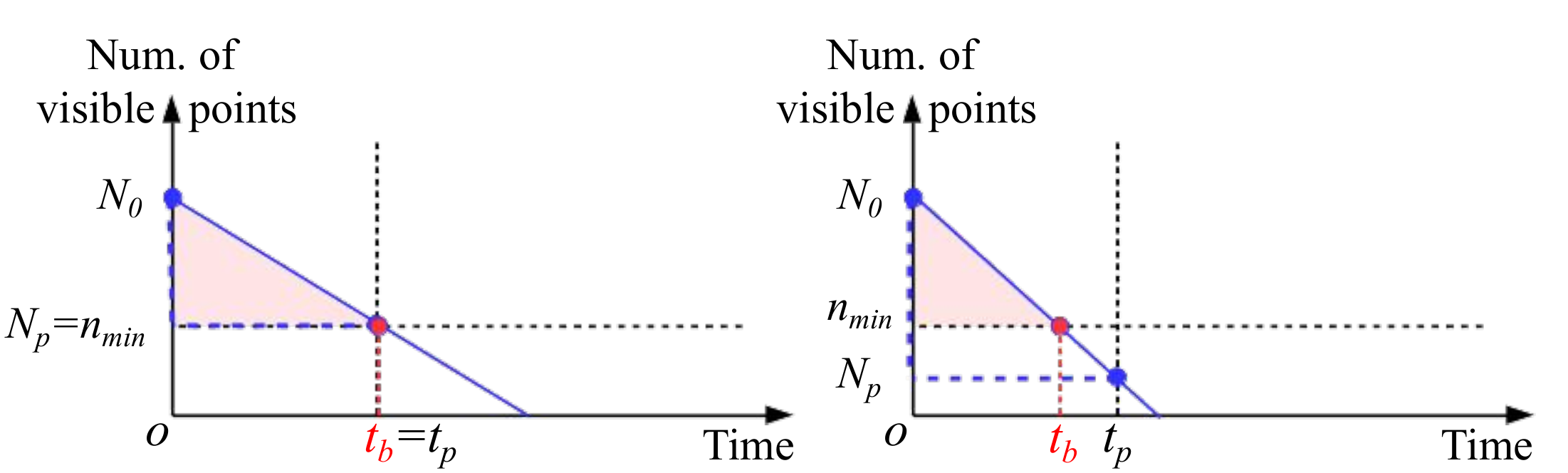}
  \caption{Linear prediction of local BA budget, where the camera pose in the near-future, $\CameraFuture$, 
  is assumed available from controller or IMU propagation.  
  The number of remaining-visible local map points $N_{p}$ is decisive in BA budget prediction.  
  Given the relative order between $N_{p}$ and a pre-set minimal number $n_{min}$, three typical 
  cases are illustrated: loose-budget, well-balance and tight-budget.
  \label{fig:Budget_Pred}} 
\end{figure*}

The {\em Good Graph} algorithm boosts the effectiveness of solving both
general and local BA problems given a root camera vertex and fixed
number of iterations in the numerical optimizer.  When combined
with several sensible modifications to improve run-time performance, the
time cost to establish a better conditioned BA sub-problem from the
root camera is a fraction of the final BA solution time cost.
As with the earlier {\em Good Feature} work \cite{zhao2020gfm}, {\em
Good Graph} for SLAM will be an active optimization approach that 
is time-cost aware with the goal of optimizing the computational budget to
improve output performance metrics (here, map RMSE and subsequently pose
RMSE). In this context, active optimization means an approach that
actively selects from available data to define an optimization problem
whose limited scale solution provides low-error estimates, and with more
predictable and consistent optimization times.  What remains is to
establish a mechanism for adaptively implementing the {\em Good Graph}
algorithm within an on-board SLAM system.

Compared to more general use of BA, one key aspect of local BA within VSLAM
back-ends is the impact of optimization time on performance.  The local
BA should be solved in time to contribute to the local pose estimation
front-end process:  longer-than-expected BA processing could cause
drift or even track failure, especially for platforms under fast
movement.  Thus, online implementation requires a mechanism to bound the
problem size to meet anticipated time budget constraints.
Optimization time and BA problem size are directly related quantities.
Thus, the objective of this section is to communicate a strategy for
determining the time budget of the local BA problem, as well as the size
$k$ of the desired {\em good subgraph}. The desired size $k$ versus the
current size of the (covisilibility-limited) local BA problem will
determine whether to trigger the {\em Good Graph} selection process or not.



\subsection{Predicting Budget of Local BA}
The primary role of the local BA in VSLAM is to provide 
accurate map points as localization references for future camera frames.
When few map points will be visible in future camera frames, it is
necessary to execute local BA at a fast rate so that estimates of
new map points converge in time.  
When sufficient map points will be present in future frames, the local BA
can run at a slower rate to provide a complete and fully-optimized map.
In effect, the budget of the local BA problem should be related to the
number of visible map points in future frames.

Similar to the feature selection work \cite{carlone2019attention}, 
camera poses (with noise) are assumed to be available in the near future.  
This assumption is reasonable: for closed-loop systems such as robots, 
future poses are available from the trajectory tracking controller or
the trajectory planner.  
For open-loop systems such as AR headsets, near-future poses can be 
predicted by propagating IMU measurements.  
Given future camera poses up to $\dtfuture$
seconds from the current time, there will be a set of 
visible map points that project to the predicted camera views, whose
cardinality is $\Nfuture$.
Now, assume that this quantity of visible map points decays linearly with
time, defining a function $N:{\mathcal R}^+ \rightarrow \mathcal{N}$
where the argument is time into the future.
The local BA time budget should be such that the map point estimates are
improved prior to losing visibility for a significant fraction of them.

The value $N(0) = N_0$ is the number of points visible in the current
(key)frame, The predicted triplet of values ($N_0$, $\Nfuture$, and
$\dtfuture$) form a triangle in the number of visible features versus
time advance graph, see Fig.~\ref{fig:Budget_Pred} (blue dash-dotted
triangle).  It defines the slope of the assumed linear loss function
$N(\cdot)$.  Given an acceptable lower limit of remaining points $\Nmin$
and the current amount of visible points $N(0)$, there will be a time
point $\dtbudget$ at which $N(\dtbudget)$ decays to the minimum
acceptable quantity of visible map points $\Nmin$ based on the assumed
linear loss in visibility.  These variables ($N(0)$, $\Nmin$, and
$\dtbudget$) define a different triangle (shaded triangles), and can
have a different size relative to the predicted triplet triangle.  The
ratio of the triangle areas is an indicator of time excess or
insufficiency. It can be used to modulate up or down the local BA time
budget $\dtbudget$ based on similar triangles geometry for the case that
there is reduced map point visibility into the future:
\begin{equation} \label{eq:Budget_Pred}
  t_{b} =
    \begin{cases}
      \dfrac{N_0 - \Nmin}{N_0 - \Nfuture}t_{p} & \text{if } \Nfuture < N_0\\
      t_{max} & \text{otherwise}
    \end{cases}       
\end{equation}
If there are more map points visible in the future than at the current
time, then the local BA problem is presumed to be well-conditioned as
there are many points available. The back-end has most likely optimized
the points already such that they will sufficiently inform pose
estimation along the planned or predicted camera trajectory. There is no
need to limit the local BA budget in order to generate well conditioned
map points sooner, thus the local BA process is given a larger time 
allocation $t_{max}$. 
The main parameters were set as follows: $\dtfuture= 500$ ms, 
$t_{max} = 800$ ms, and $\Nmin = 240$ points.

The time allocation provided will be scenario dependent as the triangle
generated will fluctuate with $N_0$ and $\Nfuture$, which depend on what has
been measured and what will be measured. It is data-adaptive. 
The local BA budget predicted with \eqref{eq:Budget_Pred} implicitly
reflects structure and motion information.  
Three typical cases of local BA budget outcomes based on the linear
visibility loss model are illustrated in Fig. \ref{fig:Budget_Pred}, 
which are determined by the predicted visible points $\Nfuture$ being
larger than, equal to, or less than the preset minimum points $\Nmin$,
as seen from left to right.  
The first case maps to a larger budget since the predicted loss is not
as bad as the largest acceptable loss (which hits the lower limit). The
area of the triangle is smaller, meaning that there is extra time to
perform back-end calculations. 
It is likely to occur when the scene structure is texture-rich, as
abundant map points are visible, or when the camera motion is slow, as
most points remain visible because of the small parallax.
The third case maps to a smaller budget since the predicted loss goes
below the acceptable limit. The opposite relationship occurs, in that
the predicted triangle has more area than the largest acceptable
triangle under the comparable linear loss rate. 
It suggests reducing the time allocation in order to more rapidly
estimate the few map points that are anticipated to be visible in the
near future.  This latter case typically appears when the future world
structure has limited texture or the camera is moving rapidly.  
In the equality case, the
expected amount of visible map points, the estimation convergence rate
of trajectory visible map points, and the optimization time cost of the
back-end process are presumed to be in balance.  

\subsection{Determining the Size of Good Graph}
Given a certain budget $\dtbudget$ for the local BA problem, a
comparably sized subgraph needs to be selected for use within the 
{\em Good Graph} algorithm so that the downstream local BA solution time
fits within the target time budget.  
The key parameter to be sent into the {\em Good Graph} selection is the
desired subgraph size $k$, characterized by the number of keyframes. The
relationship between $\dtbudget$ and $k$ will vary based on the
available computation. It must be empirically derived, either beforehand
or during online operation.

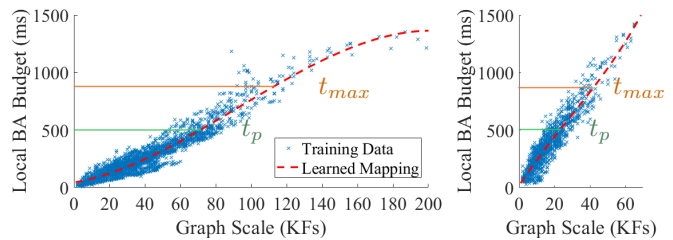
\begin{figure}[t]
  \centering
  \definecolor{darkgreen}{rgb}{0.0,0.3,0.10}
  \begin{tikzpicture}[inner sep=0pt,outer sep=0pt]
    \node (PC) at (0in,0in)
      {\includegraphics[scale=0.215]{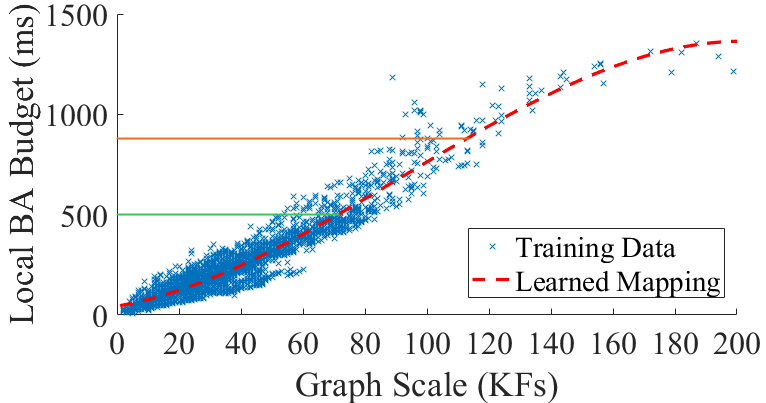}};
    \node[anchor=south west,xshift=3pt] (Jet) at (PC.south east)
      {\includegraphics[scale=0.215]{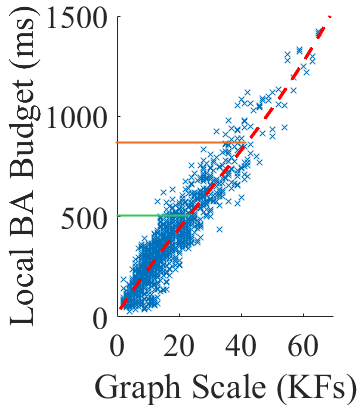}};
    \node[xshift=10pt,yshift=-3.5pt,darkgreen!70] at (PC) {$t_p$};
    \node[xshift=45pt, yshift=13pt, orange!80!black] at (PC) {{$t_{max}$}};
    \node[xshift=17pt,yshift=-3.5pt,darkgreen!70] at (Jet) {$t_p$};
    \node[xshift=32pt, yshift=13pt, orange!80!black] at (Jet) {{$t_{max}$}};
  \end{tikzpicture}
  \caption{Mappings between keyframe number and local BA budget, on two target devices.
  {\bf Left}: mapping learned on PC with Intel i7-7700K CPU.
  {\bf Right}: mapping learned on Jetson TX2 with ARM SoC (Cortex A57).
    \label{fig:Budget_Map}} 	
\end{figure}

In either case, the easiest manner to establish the necessary
relationship between $\dtbudget$ and $k$ is to regress on the function
predicting the time from the keyframe quantity, $\dtbudget = \tfun(k)$,
based on empirically measured local BA times to problems of different
sizes computed on the target device. 
Inversion of the function gives the
desired mapping $k = \kfun(\dtbudget) = \tfun^{-1}(\dtbudget)$.
Fig. \ref{fig:Budget_Map} provides two example functions used in the
experiments and the source data leading to the regressed curves. 
One regressed function is for a PC with an Intel CPU, and the other is
for an embedded device with an ARM SoC.
The dependence should be at worst cubic, therefore a cubic polynomial fit is
regressed.  During run-time, the {\em Good Graph} local BA size will be
determined by the inverse mapping $k=\kfun(\dtbudget)$ and the time
budget to meet. 
Plotted along with the curves is the nominal time budget
($\dtfuture$) to use during predicted feature loss and the maximal time
budget ($t_{max}$) to use during predicted feature increases. These map
to hardware-dependent keyframe limits.

\subsection{Integration to Local BA in VSLAM}

Given a mechanism to establish the desired size $k$ of the good
subgraph, the final step is to specify the nature of the 
{\em Good Graph} selection process in a manner compatible with the 
VSLAM system.
Earlier, Section \ref{sec:ImpDet_Valid} explored the
implementation of {\em Good Graph} from two perspectives: the
construction of the maximally sized problem (up to $k$) given all
possible camera graph vertices, and the construction of a covisibility
informed subset. It concluded that the most sensible augmentation for
VSLAM would be to sub-select from the covisible set since existing
VSLAM methods will typically already form the local BA problem based on
covisibility \cite{ORBSLAM}.  However, that leads to another
interesting data-adaptive outcome that establishes whether to trigger
{\em Good Graph} or not: the desired graph size versus the covisibility
graph size.
If the desired subgraph size $k$ is equal or larger than the size $m$ of
the candidate pool based on covisibility, {\em Good Graph} sub-selection
is skipped and the entire covisibility subgraph establishes the local
BA optimization.  If instead $k < m$, then {\em Good Graph} selection is
triggered and $k-1$ historical keyframes will be sub-selected from the
candidates.  The output good subgraph with $k$ keyframes (including the
current one) will (approximately) maximize the conditioning of the local
BA problem seeded with the current estimate, i.e., with the most recent
keyframe.

One additional modification is made to improve the overall local BA
optimization with regards to the planned or predicted future state of
the camera.  Instead of only taking the most recent keyframe as the
initial selection in Good Graph, we can further weight the selection to
those keyframes (and, implicitly, the map points) that are informative
to future estimates.  Predicting the budget of local BA requires
predicting the camera pose and the visibility of map points in the
near-future. 
The predicted camera pose should be exploited to create a local BA
problem attuned and conditioned to where the camera will be $\dtfuture$
seconds into the future. One means to do so is to create a 
virtual keyframe with the predicted map measurements at this predicted
camera pose $\CameraFuture$. 
The {\em Good Graph} selection algorithm for SLAM will seed the selection
process with both the current keyframe and the virtual keyframe at near-future.
It will then seek for the subgraph maximizing the
conditioning of current and the near-future estimates.
Once found, the subgraph without the virtual keyframe defines the local
BA problem to solve, whose size is compatible to the targeted time
budget and whose structure is optimized for the predicted camera
trajectory.

Based on the above implementation, the active {\em Good Graph} algorithm
modification for VSLAM has similar properties to the active {\em Good
Features} (GF) matching algorithm \cite{zhao2020gfm}. In the GF system,
before actively establishing which map points to match against the
current keyframe, there first exists a test to establish whether active
matching should be instantiated and how many matches should be sought,
followed by a fast selection mechanism. 
Adaptively enabling or disabling the {\em Good Features} component
based on the runtime or data-driven properties of the SLAM algorithm 
provided a unique approach to active matching by triggering the process
only when necessary. When triggered, the size of the subset selected was
determined by a desired feature matching quantity.
In doing so, it was one of the first active matching approaches to save
time, in the sense that the combined time cost of the active matching
and the reduced size pose optimization was lower than that of the full
size pose optimization.  Given that the {\em Good Graph} active
selection mechanism has a similar design approach, the properties
associated to the front-end processing of {\em Good Features} SLAM
implementations should hold for the back-end processing of {\em Good
Graph}, as the selection process is triggered only when the default
local BA optimization size is too large to complete within the given
budget.  When triggered, an efficient selection process downsizes the
local BA optimization to fit within the time budget without sacrificing
estimation accuracy.  Conversely, when there are no constraints, the
process will take as much time as specified permissible. The adaptive
compute design improves the properties of the SLAM pose estimation step
by providing timely and accurate map estimates. 


\begin{figure}[!tb]
  \centering
  \includegraphics[clip, trim=0cm 4cm 0cm 0cm, width=\linewidth]{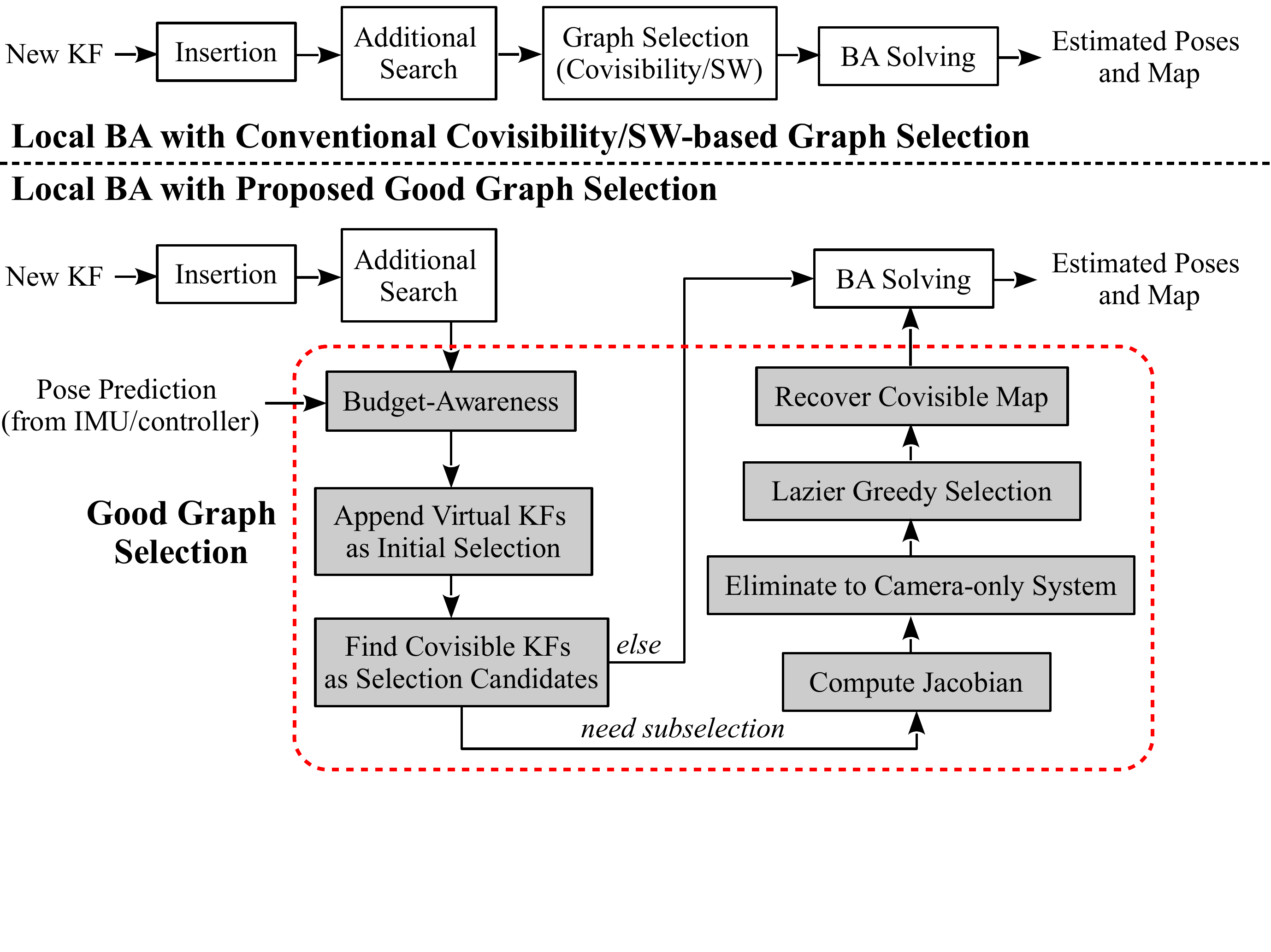}
  \caption{Pipelines of local BA in VSLAM back-end.
  \textbf{Top}: conventional local BA pipeline, where the graph to be optimized is 
  determined using heuristics such as covisibility or sliding-window (SW).
  \textbf{Bottom}: local BA pipeline with the proposed {\em Good Graph}
  selection integrated (shaded boxes), where the graph is determined
  using {\em Good Graph} algorithm when a smaller subgraph is desired
  per budget-awareness, i.e., $k<m$.
  \label{fig:GoodGraph_BA}} 
\end{figure}

A visual description for the integration of the {\em Good Graph}
algorithm into the local BA module of a VSLAM back-end is illustrated in
Fig.  \ref{fig:GoodGraph_BA}.  
Though the incremental nature is exploited in the 
{\em LogDet} computation, the overall design of {\em Good Graph}
algorithm has been implemented for batch BA, where each local BA is
configured and solved independently.  
While batch BA is commonly used in BA-based VSLAM back-ends, a back-end
with incremental BA should further improve cost-efficiency.  In the
future, a fully incremental design for {\em Good Graph} selection that
couples with incremental BA solving is worth implementing.

%
%

\section{Experimental Results} \label{sec::exp}

This section evaluates the performance of the proposed Good Graph 
algorithm in two different settings, open-loop and closed-loop.
The open-loop setting is the more common benchmarking and evaluation
approach, where the recorded data is re-played and processed. 
In this case, there
is the option of processing with or without time constraints. The former
option is chosen, which involves missed frames if the algorithm cannot
process at the rates of the recorded signal, as the same would occur in
real-world operation.
The closed-loop setting is less commonly explored due to issues with
reproducibility and ground-truth. 
Here, the use of simulation via ROS/Gazebo permits consistent and
reproducible evaluation of SLAM systems, with the outcomes valid for
settings with similar visual and dynamic characteristics 
\cite{bodin2018slambench2,ye2019characterizing,zhao2020closednav}.  
Closed-loop operation means that the SLAM pose estimate is used to
generate the feedback control signal for a trajectory tracking module.
Poor pose estimation will lead to poor trajectory tracking.  
In both evaluation settings, the Good Graph algorithm modification
improves the cost-efficiency of BA and, consequently, the system
performance.

\subsection{Good Graph in Standalone VSLAM}
Several {\em Good Graph} variants and baseline SLAM implementations
inform the benchmarking analysis.  Their descriptions and configurations
are found here along with common experimental details.

\subsubsection{Grood Graph Implementation and Variants}
The \emph{Good Graph} algorithm was integrated into 
GF-ORB-SLAM ({\em GF} \cite{zhao2020gfm}), a BA-based, stereo VSLAM
system based on ORB-SLAM.  Relative to the canonical ORB-SLAM ({\em ORB}
\cite{murORB2}), the {\em GF} modified front-end improves cost-efficiency
through active (\emph{map-to-frame}) feature matching.  
The back-ends of {\em GF} and {\em ORB} are identical: they both 
use covisibility to bound the local BA process. 
The \emph{Good Graph} algorithm was integrated into the BA-based back-end 
of {\em GF}, dubbed {\em GF+GG}. 
Implementation of the Good Graph algorithm uses SLAM++ 
\cite{ila2017slam++}, which supports block matrix manipulation 
and incremental factorization.  
To simulate access to future poses in the open-loop setting, we add 10\%
white Gaussian error to the ground truth poses, and feed the noisy pose
predictions to the budget-awareness module.  
Meanwhile, in the closed-loop setting, the budget-awareness module
directly obtains future poses from the trajectory controller.  The
mapping between the local BA time budget and the desired subgraph size
was estimated \textit{a priori}, as per Fig.~\ref{fig:Budget_Map}.  

To assess \emph{Good Graph} relative to other potential implementations, 
three GF-ORB-SLAM variants are included as evaluation baselines, 
with one of them being the standard GF-ORB-SLAM ({\em GF}).  
The second variant is a sliding-window strategy implemented for the BA
back-end of {\em GF}, leading to a combined system denoted {\em GF+SW}.  
The third variant implements an aggressive state selection strategy
based on covisibility: only the top-$k$ covisible
camera states get passed along to the local BA, while all covisible camera 
states are used in {\em GF}.  
It is referred to as {\em GF+CV}.  
For both {\em GF+SW} and {\em GF+CV} the problem size is bounded to 30
keyframes in local BA. All variants above are run as stereo only SLAM
implementations (i.e., without using IMU data).

\subsubsection{Baseline Methods}
Five state-of-the-art visual(-inertial) SLAM systems with stereo camera
support were included: 
{\em ORB} \cite{murORB2}, 
{\em SVO} \cite{SVO2017},
VINS-Fusion \cite{qin2018vins},
ICE-BA \cite{liu2018ice},
and a visual-inertial implementation of MSCKF \cite{sun2018robust}.  
{\em ORB} stands for the canonical ORB-SLAM, which is a stereo visual-only SLAM system 
with feature-based front-end and BA-based back-end.  
{\em SVO} is a lightweight, visual-only odometry system with a direct front-end 
and a sliding window BA back-end.  
By skipping explicit feature extraction and matching, {\em SVO} is computationally 
lighter than the feature-based {\em ORB} and {\em GF} variants.  
VINS-Fusion, labeled {\em VIF}, is a visual-inertial SLAM system that 
performs sliding window BA in the back-end.  
ICE-BA, labeled {\em ICE}, is an incremental and sliding window BA visual-inertial system.
The visual-inertial implementation of MSCKF is dubbed {\em MSC}, and is included to also test a filter-based VSLAM.  
All three visual-inertial systems, namely {\em VIF}, {\em ICE} and {\em
MSC}, track sparse optical flow in the front-end.  

%
%

\subsubsection{Implementation / Experimental Setup}
Across the \emph{Good Graph} variants and the baseline methods,
a variety of back-end options are covered: 
covisibility and batch BA, 
sliding window and batch BA, 
sliding window and incremental BA, 
{\em Good Graph} and batch BA, and 
filter-based.  
A parameter sweep identified configuration parameters with good
performance for these VSLAM systems.

General performance evaluation involves application of the test SLAM systems 
in {\em open-loop} with recorded sensor data. Again, {\em open-loop}
refers to the fact that the estimation data does not impact actuation
and therefore does not impact future sensor data. 
For the basic set of {\em open-loop} tests, the sensor signals were
provided at their collection rate.  This play-back rate would reflect
SLAM estimation under normal conditions for the hardware used.

To assess the performance of VSLAM system under different computational
limits, the same sensing data was provided at a higher frequency given by a
multiple of the collection frequency, which we call \textit{fast-mo}.  
This includes dropping or ignoring sensing data if the SLAM process has
not yet completed from the prior sensor input cycle, in order to
incorporate hard real-time constraints. 
Inspired by the idea of {\em slo-mo} in VSLAM benchmarking
\cite{ye2019characterizing}, {\em fast-mo} evaluation attempts to
simulate different levels of computational limits and avoids the need to 
configure the evaluated VSLAM systems across multiple devices with
different computational resources. 
In {\em fast-mo}, the VSLAM systems are configured on a single computer 
(Intel i7-7700K CPU, PassMark 2583 per core), 
but are evaluated using different rates of visual input data.  
VSLAM performance on a PC with 4x {\em fast-mo} give a rough upper
bound of a method's actual real-time performance on a 4-time slower device 
(with less cache, lower data throughput rate, lower transmission rate, etc.).  
Five levels of {\em fast-mo} speeds are evaluated, ranging from 1x to 5x.  
A low-power CPU can be simulated with 2x and 3x {\em fast-mo}, 
while an ARM SoC can be simulated with 4x and 5x {\em fast-mo}  
\cite{blem2013detailed}. 

Since lower powered compute hardware is often enhanced by additional
hardware-acceleration modules, an additional set of experiments was
performed for the variants to simulate the effect of working with a
hardware-accelerated front-end. 
This test set assesses the 
performance impact of the {\em Good Graph} local BA back-end when combined with
co-computing modules. Rather than providing the raw video data, the
implementations are given precomputed feature points.
Doing so removes the overhead of the front-end feature extraction 
(i.e., from 16 ms to 3 ms per stereo frame).  
Performance of an actual VSLAM system consisting of hardware
acceleration with {\em GF}/{\em GG} algorithmic improvements should fall 
between the regular {\em fast-mo} results and the precomputed results.  


%
\begin{figure*}[t]
  \vspace*{0.5em}
  \includegraphics[width=0.34\linewidth]{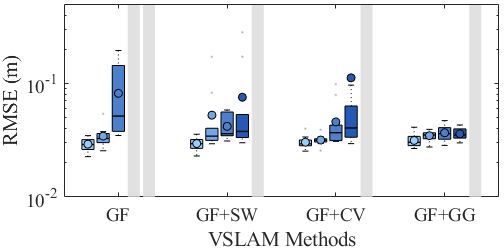} 
  \includegraphics[width=0.325\linewidth]{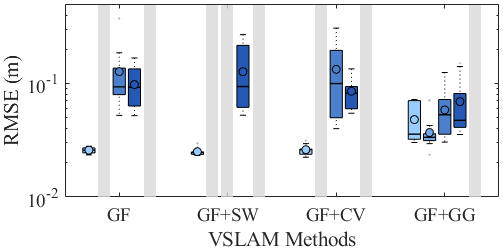} 
  \includegraphics[width=0.325\linewidth]{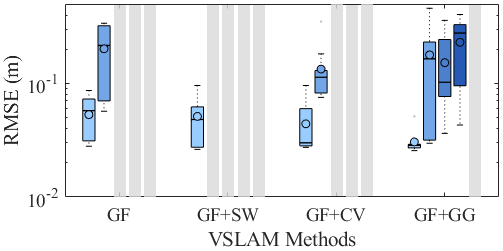}\\
  \vspace*{-0.2in}
  \caption{{\em Fast-mo} results of GF variants without precomputed keypoints, on three EuRoC sequences: 
  {\em MH 03 med} ({\bf left}), {\em V1 02 med} ({\bf middle}) and {\em V1 03 diff} ({\bf right}). 
See text for plot details. {\em GF+GG} shows advantage over other GF variants in terms of both track success and RMSE.
    \label{fig:EuRoC_RMSE_Onl}} 	

  \vspace*{0.5em}
  \includegraphics[width=0.34\linewidth]{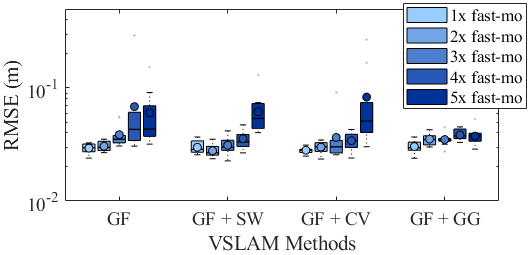} 
  \includegraphics[width=0.325\linewidth]{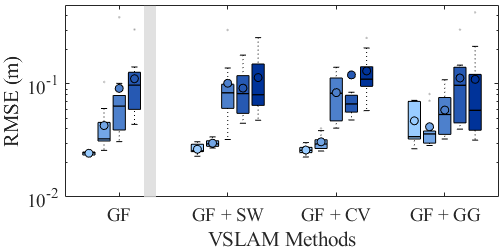} 
  \includegraphics[width=0.325\linewidth]{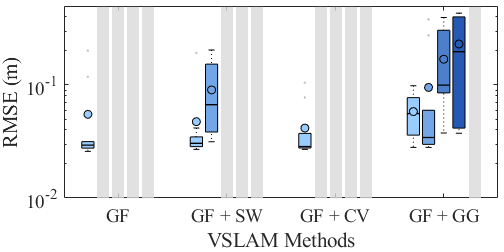}\\
  \vspace*{-0.2in}
  \caption{{\em Fast-mo} results of GF variants with precomputed keypoints, on the same three sequences.  
  Compared with Fig.~\ref{fig:EuRoC_RMSE_Onl}, track failures happened
  less because of the removal of the front-end bottleneck.  The {\em GF+GG} 
  variant addresses the cost-efficiency issue of BA-based back-end
  nicely, as it tracks on more {\em fast-mo} settings (especially $\geq$
  3x) than the rest, while keeping lower or similar RMSE.
  \label{fig:EuRoC_RMSE_Pre}} 	
  
  \vspace*{0.5em}
  {\includegraphics[width=\columnwidth,clip=true,trim=0in 3.65in 0in 0in]{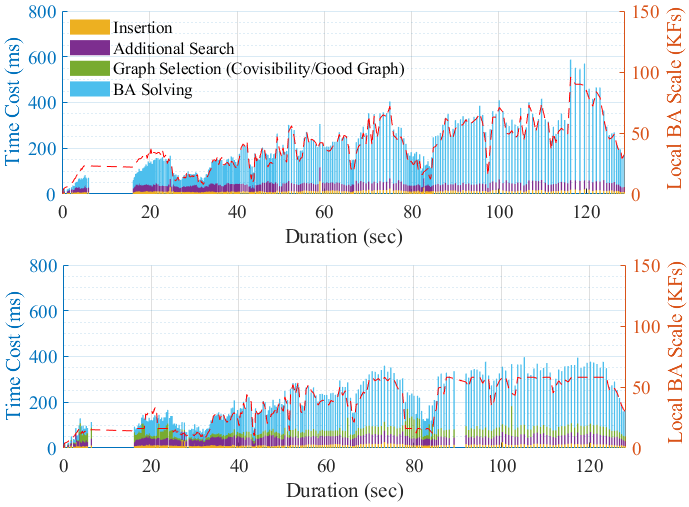}}
  \hfill
  {\includegraphics[width=\columnwidth,clip=true,trim=0in 0.10in 0in 3.55in]{EuRoC/GoodGraph_bar_MH_03_medium_x2_v2.png}}
  \caption{Time cost breakdowns (left $y$-axis) and graph size (right
  $y$-axis) of local BA specification and optimization on {\em MH 03 med}, 
  at 2x {\em fast-mo}.
  No local BA is triggered between 7 sec and 15 sec since the camera
  remains static.  
  {\bf Left}: {\em GF} uses all covisible keyframes in the local BA
  based on the {\em ORB} implementation.
  {\bf Right}: {\em GF+GG} selects and optimizes the good subgraph in
  local BA. \label{fig:TimeProfile}} 	
\end{figure*} 

\subsubsection{Dataset and Evaluation Criteria}
The open-loop evaluation is based on the EuRoC 
benchmark \cite{burri2016euroc}, which contains 11
stereo-inertial sequences recorded in three different indoor
environments.  
The performance of each VSLAM system is evaluated using the real-time pose 
tracking output.  
Real-time pose tracking accuracy is measured by the \textit{absolute
root-mean-square error} (RMSE) \cite{sturm12iros_ws}
between the ground truth trajectory and the real-time VSLAM output from
the pose estimation step, after conducting an {\em SE3} alignment (with
no scale correction).  Robustness of each VSLAM system is quantified by 
the number of successfully tracked sequences versus the total number of sequences. 
For each configuration (benchmark sequence, VSLAM system, {\em fast-mo} speed, 
with/without precomputing), a 10-run repeat was executed.  
Track failure is concluded when more than 40\% of frames are not tracked 
successfully.  Results were reported only if zero tracking failure 
occurred during the 10-run repeat.  Otherwise, the results of the 
corresponding configuration were discarded since the VSLAM system output 
was not reliable.

Table~\ref{tab:EuRoC_Stereo_RMSE} presents the open-loop {\em fast-mo} 
outcomes of different VSLAM methods on the EuRoC benchmark. 
For each sequence, the lowest RMSE values with the same category 
(defined by {\em fast-mo} speed and with/without precomputing) are highlighted in bold.  
The average RMSE values of each VSLAM method, across the columns, are
listed in the rightmost column, where those with more track loss than
{\em GF+GG} are marked in parentheses.  Table~\ref{tab:EuRoC_Stereo_RMSE}
is the source data for the analysis presented in the next two subsections.

\begingroup
\setlength{\tabcolsep}{5pt} 
\renewcommand*\arraystretch{0.975}
\begin{table*}[!tb]
	\footnotesize
	\centering
	\caption{RMSE (m) on EuRoC Stereo Sequences for Standard Computer\label{tab:EuRoC_Stereo_RMSE}}
	\begin{tabular}{|c|c|ccccccccccc|c|}
		\toprule
		\textbf{ } & \textbf{ } & 
		\multicolumn{12}{c|}{\bfseries \small Sequences} \\
		\textbf{Cfg} & \textbf{Methods} & \textit{MH 01} & \textit{MH 02} & \textit{MH 03} & \textit{MH 04} & \textit{MH 05} & \textit{V1 01} & \textit{V1 02} & \textit{V1 03} & \textit{V2 01} & \textit{V2 02} & \textit{V2 03} & \textbf{Avg.} \\
		\midrule
		\parbox[t]{1mm}{\multirow{4}{*}{\rotatebox[origin=c]{90}{1x; Pre}}}
& GF 	  &  0.025 & 0.019 & 0.029 & 0.109 & 0.064 &\bf 0.035 &\bf 0.024 & 0.055 &\bf 0.043 &\bf 0.040 & 0.228 & 0.061 \\ 
& GF + SW &  0.023 &\bf 0.018 & 0.030 &\bf 0.104 & 0.064 & 0.036 & 0.026 & 0.047 & 0.046 &\bf 0.040 & 0.183 & 0.056 \\ 
& GF + CV &\bf 0.021 &\bf 0.018 &\bf 0.028 & 0.113 &\bf 0.061 & 0.036 & 0.026 &\bf 0.042 & 0.046 & 0.042 &\bf 0.160 &\bf 0.054 \\ 
& GF + GG &  0.024 &\bf 0.018 & 0.030 & 0.110 & 0.070 & 0.036 & 0.047 & 0.058 &\bf 0.043 & 0.046 & 0.173 & 0.060 \\ 
		\midrule
		\parbox[t]{1mm}{\multirow{9}{*}{\rotatebox[origin=c]{90}{1x}}}
& GF 	&  0.023 &\bf 0.018 &\bf 0.029 & 0.119 & 0.066 &\bf 0.036 & 0.026 & 0.053 &\bf 0.042 & 0.046 & 0.186 & 0.059 \\ 
& GF+SW &  0.023 &\bf 0.018 &\bf 0.029 &\bf 0.094 &\bf 0.062 &\bf 0.036 &\bf 0.025 & 0.051 & 0.046 & 0.045 & 0.220 & 0.059 \\ 
& GF+CV &\bf  0.021 & 0.019 & 0.030 & 0.115 & 0.065 &\bf 0.036 & 0.026 & 0.044 & 0.046 & 0.046 & 0.194 & 0.058 \\ 
& GF+GG &  0.027 & 0.019 & 0.031 & 0.107 & 0.073 &\bf 0.036 & 0.048 &\bf
0.030 &\bf 0.042 &\bf 0.036 &\bf 0.161 &\bf 0.056
\\\cline{2-14}\rule{0pt}{9pt}
& ORB &  0.030 & 0.025 & 0.032 & 0.166 & 0.089 & 0.037 & 0.034 & 0.310 & 0.049 & 0.083 & 0.304 & 0.105 \\ 
& MSC &  - & 0.139 & 0.239 & 0.273 & 0.245 & 0.092 & 0.105 & 0.168 & 0.063 & 0.136 & 1.290 & (0.275) \\ 
& ICE &  0.115 & 0.075 & 0.119 & 0.233 & 0.237 & 0.066 & 0.096 & 0.153 & 0.115 & 0.101 & 0.175 & 0.135 \\ 
& VIF &  1.152 & 1.324 & 1.490 & 2.243 & 2.310 & 0.345 & 0.350 & 0.533 & 0.474 & 0.373 & 0.268 & 0.987 \\ 
& SVO &  0.126 & 0.102 & 0.242 & 1.942 & 0.292 & 0.092 & 0.269 & 0.359 & 0.159 & 0.236 & 2.471 & 0.572 \\ 
		\midrule
		\parbox[t]{1mm}{\multirow{4}{*}{\rotatebox[origin=c]{90}{2x; Pre}}}
& GF &  0.025 &\bf 0.019 & 0.030 &\bf 0.107 &\bf 0.064 & 0.060 & 0.043 & - & 0.045 & 0.064 &\bf 0.144 & (0.060) \\ 
& GF+SW &  0.028 &\bf 0.019 &\bf 0.028 & 0.132 & 0.067 &\bf 0.036 &\bf 0.030 &\bf 0.090 &\bf 0.042 & 0.057 & 0.206 & 0.067 \\ 
& GF+CV &  0.028 & 0.020 & 0.030 & 0.125 & 0.069 & 0.037 & 0.031 & - & 0.043 & 0.052 & 0.167 & (0.060) \\ 
& GF+GG &\bf 0.024 & 0.020 & 0.035 & 0.138 & 0.079 &\bf 0.036 & 0.042 & 0.095 & 0.043 &\bf 0.049 & 0.150 &\bf 0.065 \\ 
		\midrule
		\parbox[t]{1mm}{\multirow{9}{*}{\rotatebox[origin=c]{90}{2x}}}
& GF  &  0.028 &\bf 0.021 & 0.034 & 0.133 &\bf 0.083 & 0.037 & - & 0.202 & 0.044 & 0.071 &\bf 0.113 & (0.077) \\ 
& GF+SW &  0.030 &\bf 0.021 & 0.052 &\bf 0.108 & 0.085 &\bf 0.036 & - & - & 0.046 & 0.068 & 0.117 & (0.062) \\ 
& GF+CV &  0.025 & 0.022 &\bf 0.032 & - & 0.099 &\bf 0.036 & - &\bf 0.134 & 0.048 & 0.058 & 0.123 & (0.064) \\ 
& GF+GG &\bf 0.024 & 0.022 & 0.035 & 0.118 & 0.084 &\bf 0.036 &\bf 0.037
& 0.178 &\bf 0.041 &\bf 0.051 & 0.194 &\bf 0.075 \\
\cline{2-14}\rule{0pt}{9pt}
& ORB &  0.032 & 0.026 & 0.059 & 0.190 & 0.097 & 0.037 & - & - & - & 0.174 & - & (0.088) \\ 
& MSC &  - & 0.169 & 0.228 & 0.244 & 0.283 & 0.102 & 0.127 & 0.189 & 0.077 & 0.150 & - & (0.174) \\ 
& ICE &  0.116 & 0.074 & 0.147 & 0.276 & 0.249 & 0.062 & 0.094 & 0.146 & 0.113 & 0.102 & 0.299 & 0.152 \\ 
& VIF &  1.153 & 1.324 & 1.505 & 2.255 & 2.316 & 0.345 & 0.350 & 0.552 & 0.467 & 0.374 & 0.264 & 0.991 \\
& SVO &  0.127 & 0.106 & 0.220 & 2.081 & 0.438 & 0.097 & 0.292 & 0.409 & 0.186 & 0.260 & 2.391 & 0.601 \\ 
		\midrule
		\parbox[t]{1mm}{\multirow{4}{*}{\rotatebox[origin=c]{90}{3x; Pre}}}
& GF &  0.033 &\bf 0.021 & 0.039 & 0.155 & 0.083 & 0.038 & 0.091 & - & 0.043 & 0.088 &\bf 0.158 &\bf 0.075 \\ 
& GF+SW &  0.029 & 0.022 &\bf 0.031 &\bf 0.138 & 0.083 & 0.045 & 0.102 & - &\bf 0.042 & - & 0.195 & (0.076) \\ 
& GF+CV &  0.026 & 0.023 & 0.036 & 0.145 &\bf 0.072 & 0.052 & 0.084 & - & 0.045 & - & 0.159 & (0.071) \\ 
& GF+GG &\bf 0.023 &\bf 0.021 & 0.035 & 0.156 & 0.087 &\bf 0.036 &\bf 0.059 &\bf 0.169 &\bf 0.042 &\bf 0.059 & 0.198 & 0.080 \\ 
		\midrule
		\parbox[t]{1mm}{\multirow{9}{*}{\rotatebox[origin=c]{90}{3x}}}
& GF &  0.031 & 0.024 & 0.082 & 0.174 &\bf 0.090 & 0.040 & 0.127 & - &\bf 0.048 & 0.093 & - & 0.079 \\ 
& GF+SW &  0.033 &\bf 0.022 & 0.042 &\bf 0.151 & 0.099 & 0.038 & - & - & 0.050 & - &\bf 0.153 & (0.073) \\ 
& GF+CV &  0.035 & 0.027 & 0.046 & 0.160 & 0.095 & 0.039 & 0.133 & - & 0.061 & 0.094 & - &\bf 0.077 \\  
& GF+GG &\bf 0.027 & 0.023 &\bf 0.037 & 0.154 & 0.098 &\bf 0.037 &\bf
0.059 &\bf 0.152 &\bf 0.048 &\bf 0.059 & 0.169 & 0.078 \\
\cline{2-14}\rule{0pt}{9pt}
& ORB &  0.029 & 0.026 & 0.047 & - & 0.095 & - & - & - & 0.131 & - & - & (0.066) \\ 
& MSC &  - & 0.151 & 0.274 & - & 0.329 & 0.108 & 0.136 & 0.230 & 0.078 & 0.158 & - & (0.183) \\ 
& ICE &  0.112 & 0.261 & 0.420 & 0.236 & 0.232 & 0.066 & 0.094 & 0.190 & 0.121 & - & - & 0.192 \\ 
& VIF &  1.029 & 1.042 & 1.476 & 2.273 & 2.313 & 0.320 & 0.347 & 0.549 & 0.434 & 0.365 & 0.262 & 0.946 \\ 
& SVO &  0.117 & 0.117 & 0.282 & 1.872 & 0.447 & 0.094 & 0.293 & 0.339 & 0.220 & 0.399 & 2.514 & 0.609 \\ 
		\midrule
		\parbox[t]{1mm}{\multirow{4}{*}{\rotatebox[origin=c]{90}{4x; Pre}}}
& GF &  0.035 & 0.028 & 0.068 &\bf 0.147 & 0.088 & 0.050 & 0.111 & - &\bf 0.043 & - & 0.202 & (0.086) \\ 
& GF+SW &  0.037 & 0.025 & 0.035 & 0.166 & 0.089 & 0.039 &\bf 0.092 & - & 0.048 & - &\bf 0.144 & (0.075) \\ 
& GF+CV &  0.032 &\bf 0.023 &\bf 0.034 & 0.182 & 0.103 & 0.039 & 0.120 & - &\bf 0.043 & - & - & (0.072) \\ 
& GF+GG &\bf 0.027 &\bf 0.023 & 0.038 & 0.161 &\bf 0.082 &\bf 0.036 & 0.113 &\bf 0.231 & 0.045 &\bf 0.063 & 0.202 &\bf 0.093 \\ 
		\midrule
		\parbox[t]{1mm}{\multirow{9}{*}{\rotatebox[origin=c]{90}{4x}}}
& GF  &  - &\bf 0.022 & - & 0.189 & 0.123 & - & 0.098 & - & 0.065 & - & - & (0.100) \\ 
& GF+SW & - &\bf 0.022 & 0.075 &\bf 0.178 & 0.142 & - & 0.127 & - & 0.113 & 0.145 & - & 0.136 \\ 
& GF+CV & - & 0.023 & 0.112 & 0.201 & 0.115 & - & 0.085 & - & 0.081 & - & - & (0.103) \\  
& GF+GG & - & 0.024 &\bf 0.036 &\bf 0.178 &\bf 0.095 & - &\bf 0.069 &\bf
0.230 &\bf 0.047 &\bf 0.064 & - &\bf 0.093 \\ \cline{2-14}\rule{0pt}{9pt}
& ORB &\bf 0.031 & 0.024 & - & 0.210 & 0.118 &\bf 0.039 & 0.105 & - & 0.210 & 0.101 & - & 0.105 \\ 
& MSC &  - & 0.223 & 0.262 & 1.319 & 0.317 & 0.149 & 0.162 & - & 0.096 & 0.150 & - & 0.335 \\ 
& ICE &  0.096 & 0.808 & - & 0.258 & 0.225 & 0.070 & - & - & 0.113 & - & - & (0.262) \\ 
& VIF &  - & - & - & - & - & - & - & - & - & - & - & - \\ 
& SVO &  0.122 & 0.114 & 0.457 & 2.173 & 0.400 & 0.093 & 0.445 & 0.485 & 0.183 & 0.493 & 2.311 & 0.661 \\ 
		\midrule
		\parbox[t]{1mm}{\multirow{4}{*}{\rotatebox[origin=c]{90}{5x; Pre}}}
& GF 	&  0.039 & 0.025 & 0.060 & 0.181 & 0.104 & 0.051 & - & - & 0.078 & 0.112 &\bf 0.200 & 0.094 \\ 
& GF+SW &  0.046 &\bf 0.024 & 0.062 & 0.182 & 0.117 & 0.043 & 0.114 & - &\bf 0.048 & 0.095 & 0.206 & 0.094 \\ 
& GF+CV &  0.048 & 0.025 & 0.083 & 0.202 & 0.112 & 0.048 & 0.130 & - & 0.067 & 0.132 & - 	& 0.094 \\ 
& GF+GG &\bf 0.028 & 0.026 &\bf 0.037 &\bf 0.173 &\bf 0.100 &\bf 0.038 &\bf 0.110 & - &\bf 0.048 &\bf 0.067 & - &\bf 0.070 \\ 
		\midrule
		\parbox[t]{1mm}{\multirow{9}{*}{\rotatebox[origin=c]{90}{5x}}}
& GF  	& - & - & - & - & - & - & - & - & - & - & - & - \\ 
& GF+SW & - & - & - & - & - & - & - & - & - & - & - & - \\ 
& GF+CV & - & - & - & - & - & - & - & - & - & - & - & - \\  
& GF+GG & - & - & - & - & - & - & - & - & - & - & - & - \\ \cline{2-14}\rule{0pt}{9pt}
& ORB 	& - & - & - & - & - & - & - & - & - & - & - & - \\ 
& MSC 	& - & - &\bf 0.361 & - & 0.817 & 0.129 &\bf 0.178 & - & - &\bf 0.259 &\bf 1.341 &\bf 0.514 \\ 
& ICE 	& - & - & - & - & - & - & - & - & - & - & - & - \\ 
& VIF 	& - & - & - & - & - & - & - & - & - & - & - & - \\ 
& SVO 	&\bf 0.106 &\bf 0.112 & 0.446 &\bf 1.883 &\bf 0.355 &\bf 0.094 & 0.524 &\bf 0.475 &\bf 0.216 & 0.518 & 2.363 & 0.645
        \color{black} \\
		\bottomrule	
	\end{tabular} 
\end{table*}
\endgroup

\subsection{Analysis of Good Graph Variants}

The {\em GF} method and variants all performed well at the standard (1x)
rate. On average, the back-end modifications preserved performance as
determined by robustness (all methods tracked all sequences) and
accuracy, with the {\em GF+GG} RMSE being equal to or lower than (within
5\%) the {\em GF} version, as indicated by the last column in
Table~\ref{tab:EuRoC_Stereo_RMSE}.  In these cases, there is usually
sufficient time for the local BA process to influence pose estimation.  
As the {\em fast-mo} speed increases, the time allocation of the SLAM 
computation shrinks relative to the time between camera measurements, 
and the performance outcomes diverge. 
The {\em GF}, {\em GF+SW}, and {\em GF+CV} versions experience
degraded robustness and cannot reliably track all sequences. At 4x {\em
fast-mo} the {\em GF+GG} version starts to present reduced robustness 
(3 out of the 11 sequences has track failure). 
None of the methods track at 5x {\em fast-mo}. 
The ability of {\em GF+GG} to track through to 3x with no track loss
shows that accelerating the back-end through prioritized and time-aware
optimizations can enhance pose estimation through better optimized
map points. With precomputed keypoints, {\em GF+GG} could be pushed to 
4x rates with no track loss, and only fails on 2 sequences at 5x:Pre.

The {\em Good Graph} modification not only improves the overall
outcomes, but it exhibits lower performance variation as a function of
the \textit{fast-mo} factor.  
Fig.~\ref{fig:EuRoC_RMSE_Onl} includes boxplots of the tracking accuracy 
outcomes on three EuRoC sequences. 
In each boxplot, the outcomes are grouped by {\em GF} variant methods; 
for each variant the boxes are ordered per {\em fast-mo} speed.  
The gray columns stand for configurations with track loss.  
The {\em GF+GG} version has more consistent outcomes on {\em MH 03} and
{\em V1 02 med} (first 2 columns). Though the variation of tracking accuracy 
increases for {\em V1 03 diff}, it has the most success across the {\em fast-mo} speeds. 
None of the other variants can track beyond 2x {\em fast-mo} for
{\em V1 03 diff}.  
To factor out the front-end bottleneck in {\em fast-mo} evaluation, 
boxplots of outcomes with precomputed keypoints are provided in 
Fig.~\ref{fig:EuRoC_RMSE_Pre}. 
Although track loss happened less for the {\em GF} variants on {\em MH 03} 
and {\em V1 02 med}, {\em GF+GG} stands out, with consistent tracking accuracy 
(especially on 5x {\em fast-mo}).  
Interestingly, the pre-computing doesn't improve the track success of other 
{\em GF} variants on {\em V1 03 diff}, which indicates the bottleneck 
of this sequence is on the BA-based back-end, and that it is largely
resolved with {\em GF+GG}.

\begingroup
\setlength{\tabcolsep}{5pt} 
\renewcommand*\arraystretch{0.975}
\begin{table*}[!tb]
	\footnotesize
	\centering
	\caption{RMSE (m) on EuRoC Stereo Sequences for an Embedded Device
      \label{tab:EuRoC_Stereo_RMSE_Embedded}}
	\begin{tabular}{|c|c|ccccccccccc|c|}
		\toprule
		\textbf{ } & \textbf{ } & 
		\multicolumn{12}{c|}{\bfseries \small Sequences} \\
		\textbf{Cfg} & \textbf{Methods} & \textit{MH 01} & \textit{MH 02} & \textit{MH 03} & \textit{MH 04} & \textit{MH 05} & \textit{V1 01} & \textit{V1 02} & \textit{V1 03} & \textit{V2 01} & \textit{V2 02} & \textit{V2 03} & \textbf{Avg.} \\
		\midrule
		\parbox[t]{1mm}{\multirow{4}{*}{\rotatebox[origin=c]{90}{1x: Pre}}}
& GF 	&  - & 0.055 &\bf 0.054 &\bf 0.129 & - & 0.049 & - & - & 0.049 & - & - &\bf (0.067) \\ 
& GF+SW &  - & 0.032 & 0.113 & - & - & 0.044 & - & - & 0.063 & 0.105 & - & (0.071) \\ 
& GF+CV &  - & - & 0.057 & 0.174 & - & 0.045 &\bf 0.112 & - & 0.052 & 0.097 & - & (0.089) \\ 
& GF+GG &\bf  0.036 &\bf 0.025 & 0.061 & 0.141 &\bf 0.097 &\bf 0.038 & 0.122 & - &\bf 0.042 &\bf 0.092 & - & 0.077 \\  
		\midrule
		\parbox[t]{1mm}{\multirow{7}{*}{\rotatebox[origin=c]{90}{1x}}}
& GF 	&  - & - & - & - & - & 0.048 & - & - & - & - & - &\bf (0.048) \\ 
& GF+SW &  - & - & - & - & - & 0.050 & - & - & - & - & - & (0.050) \\ 
& GF+CV &  - & - & - & - & - & 0.066 & - & - & - & - & - & (0.066) \\ 
& GF+GG &\bf  0.037 & - &\bf 0.059 &\bf 0.176 &\bf 0.140 &\bf 0.039 & - & - &\bf 0.048 &\bf 0.084 & - & 0.083 \\\cline{2-14}\rule{0pt}{9pt}
& ORB 	&  - & - & - & - & - & - & - & - & - & - & - & - \\ 
& MSC 	&  - &\bf 0.136 & 0.249 & 0.228 & 0.355 & 0.087 &\bf 0.105 &\bf 0.179 & 0.065 & 0.119 &\bf 0.970 & 0.249 \\ 
& ICE 	&  0.100 & 0.320 & - & 0.247 & 0.224 & 0.067 & 0.107 & - & - & - & - & (0.177) \\ 
		\bottomrule	
	\end{tabular} 
\end{table*}
\endgroup

\begingroup
\renewcommand*\arraystretch{0.97}
\begin{figure*}[t]
  \includegraphics[width=0.35\linewidth]{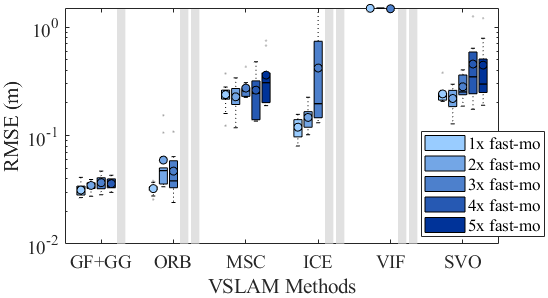}  
  \includegraphics[width=0.32\linewidth]{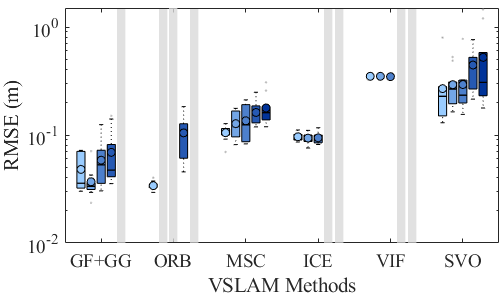} 
  \includegraphics[width=0.32\linewidth]{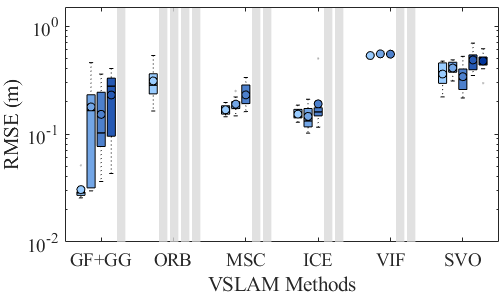}\\
  \vspace*{-0.2in}
  \caption{{\em Fast-mo} results of {\em GF+GG} and state-of-the-art
  visual(-inertial) SLAM baselines, on the same three sequences 
  ({\em MH 03, V1 02, V103}).\label{fig:EuRoC_RMSE_Final}} 	
\end{figure*}
\endgroup

To demonstrate the functionality of {\em Good Graph} in practice, 
Fig.~\ref{fig:TimeProfile} provides a time cost breakdown versus time of the
BA back-end components for a 2x {\em fast-mo} instance.  
The red curve corresponds to the right-hand $y$-axis in the figure,
which is the target local BA size versus time. 
The {\em GF} version does not modify the {\em ORB} local BA
calculations, thus they vary as a function of the covisible keyframes,
which adds variability to the local BA solver time. 
The {\em GF+GG} version bounds the local BA solver time by adaptively
choosing the quantity of keyframes used in the local BA optimization.  
During the first 60 seconds of exploration both methods have similar
time cost profiles. Afterwards, the time cost distribution of {\em GF+GG} 
is $265.6 \pm 77.5$ ms versus $259.8 \pm 95.6$ ms for the {\em GF} baseline.
These statistics are consistent with the earlier BA-only evaluation of
{\em Good Graph}. It tends to select a larger local BA problem in terms
of map points for the given target set of camera poses (as determined by
an upper limit on compute time).  Achievement of the compute bound is
indicated by the lower variance (77.5 vs 95.6).  Without this component,
the local BA problem size and its numerical solution time fluctuates
more. Further confirmation arises from examining the $L_2$ norm of the
derivative of the two curves (the local BA problem size and compute time
versus frame). The {\em GF} baseline has norms 164 keyframes/sec and 
0.8 sec/sec, while {\em GF+GG} has norms 118 keyframes/sec and 0.6
sec/sec. The latter two quantities are lower than {\em GF} values 
by 27-28\%, indicating less fluctuation of the constructed local BA
problem versus time.




\subsection{Analysis Relative to Baseline Methods}

According to Table~\ref{tab:EuRoC_Stereo_RMSE}, 
the two visual-inertial systems with sliding window BA ({\em ICE} and
{\em VIF}) had significantly higher RMSEs for 1-3x {\em fast-mo}
factors, and failed to track since 4x {\em fast-mo}.  
The two lightweight systems, direct {\em SVO} and filter-based {\em
MSC}, were robust under all five {\em fast-mo} speeds but with high RMSEs.  
The {\em GF+GG} system consistently had one of the lowest RMSEs
with full tracking success for 1-3x {\em fast-mo}. 
The track failure of {GF+GG} running on 4-5x {\em fast-mo} is partially
due to the front-end bottleneck, as seen by the full tracking at 4x:Pre
and 9 of 11 tracking at 5x:Pre. With precomputing, the error rates of 
{\em GF+GG} are at least 7 times lower than SVO.


To visualize performance progression as a function of \textit{fast-mo}
settings, Fig.~\ref{fig:EuRoC_RMSE_Final} depicts box plots of the
tracking accuracy outcomes for {\em GF+GG} and the baselines. 
The proposed {\em GF+GG} has the best average tracking accuracy in all
except for 2 cases: 
for 1x on {\em V1 02 med} where it is slightly worse than {\em ORB}, and 
for 2x on {\em V1 03 diff} where it's slightly worse than {\em ICE}.
In both cases the differences are negligible relative to the range of
RMSE values across all of the methods.
Moreover, {\em GF+GG} is the only BA-based VSLAM that tracks from 1x to
4x {\em fast-mo} on all 3 example sequences.  

This study shows that by adding a compute aware local BA back-end to the 
accelerated front-end {\em GF} of ORB-SLAM further improves
the run-time performance properties of SLAM system. {\em GF} demonstrated 
improved performance relative to published baselines \cite{zhao2020gfm}, 
while {\em GG} provides additional robustness to computational resources 
and/or timing constraints.
The performance degradation of {\em GF+GG} is relatively
graceful from 1x {\em fast-mo} to 4x {\em fast-mo}.  


\subsection{Good Graph on Embedded Device}

To confirm that the predicted outcomes of {\em fast-mo} (especially 4x 
and 5x) agree with the actual results on the equivalent low-power device, 
we ran the experiments using EuRoC sequences on an embedded device, 
Nvidia Jetson TX2 with an ARM SoC (Cortex A57) and 7.5W of power consumption.  
Evaluation results of {\em GF+GG}, alongside with three {\em GF}
variants and three baseline methods ({\em ORB}, {\em MSC}, and {\em
ICE}), are summarized at Table~\ref{tab:EuRoC_Stereo_RMSE_Embedded}.
Due to hardware compatibility issues, the {\em VIF} and {\em SVO}
baselines were dropped.

In terms of robustness and low-cost, the light-weight visual-inertial 
{\em MSC} stands out with only one tracking failure.  The integration of 
inertial information assists the data association component.  Nevertheless, 
the RMSE achieved by {\em MSC} was high, mostly due to the drift 
accumulation of the filter with a short time sliding window.  
The visual-only {\em GF+GG} is not as robust as {\em MSC}, but does 
exhibit a lower RMSE than both visual-inertial systems {\em MSC} and {\em ICE}.  
When compared to the {\em GF} variants, {\em GF+GG} shows improved
robustness as it tracked 7 of the sequences, versus 1 for the variants.
Performance of the embedded {\em GF+GG} is closest to the 4x {\em
fast-mo} case.

Using precomputed feature points, {\em GF+GG} tracks 9 of the sequences
and has a slightly lower RMSE (by 7\%). The outcomes are similar to
{\em GF+GG} for the 5x:Pre {\em fast-mo} case. The variants track more than
their 1x counterparts, but still less than {\em GF+GG}.
The clear performance improvement from {\em GF} to {\em GF+GG} suggests
that the {\em Good Graph} modification compensates for BA back-end
bottlenecks by directing available computational resources towards BA
sub-graph components helpful to the front-end.
Additional parameter tuning with an FPGA-based front-end, and possibly
the incorporation of inertial signals, should resolve the four failure
cases in {\em GF+GG} 1x (or the two in 1x:Pre), and provide a low-drift
SLAM system for small scale robotic systems or localization and mapping
devices.

\subsection{Good Graph in VSLAM-based Closed-loop Navigation \label{secExpCL}}

In practical applications such as closed-loop navigation, VSLAM
would be a subsystem within an autonomous navigation system.  
The envisioned role of VSLAM for mobile robots is to provide pose
estimates when navigating through environments.  Especially in cases
where GPS positioning is less reliable or unavailable.  
The VSLAM subsystem provides robot pose estimates to other autonomy
components, which require pose for feedback or for processing
measurements. Consequently, closed-loop navigation performance will be
affected by VSLAM subsystem performance relative to robustness,
accuracy, and latency.  This section's investigation will show that
enhancing VSLAM-based state estimation with the {\em Good Graph}
algorithm improves navigation performance.  The combination of the 
{\em Good Graph} back-end and the low-latency {\em Good Feature}
front-end, {\em GF+GG}, will be evaluated. The {\em Good Graph}
modification improves the cost-efficiency of local BA, and increases the
quality of map points available for pose tracking.  The budget-awareness
module of {\em Good Graph} uses the trajectory to track as predicted future
poses, with which it establishes a time compatible local BA budget.  

\subsubsection{Closed-loop Navigation System Design}

\begin{figure}[!tb]
  \centering
  \includegraphics[width=\columnwidth,clip=yes,trim=0in 4in 1.5in 0in]{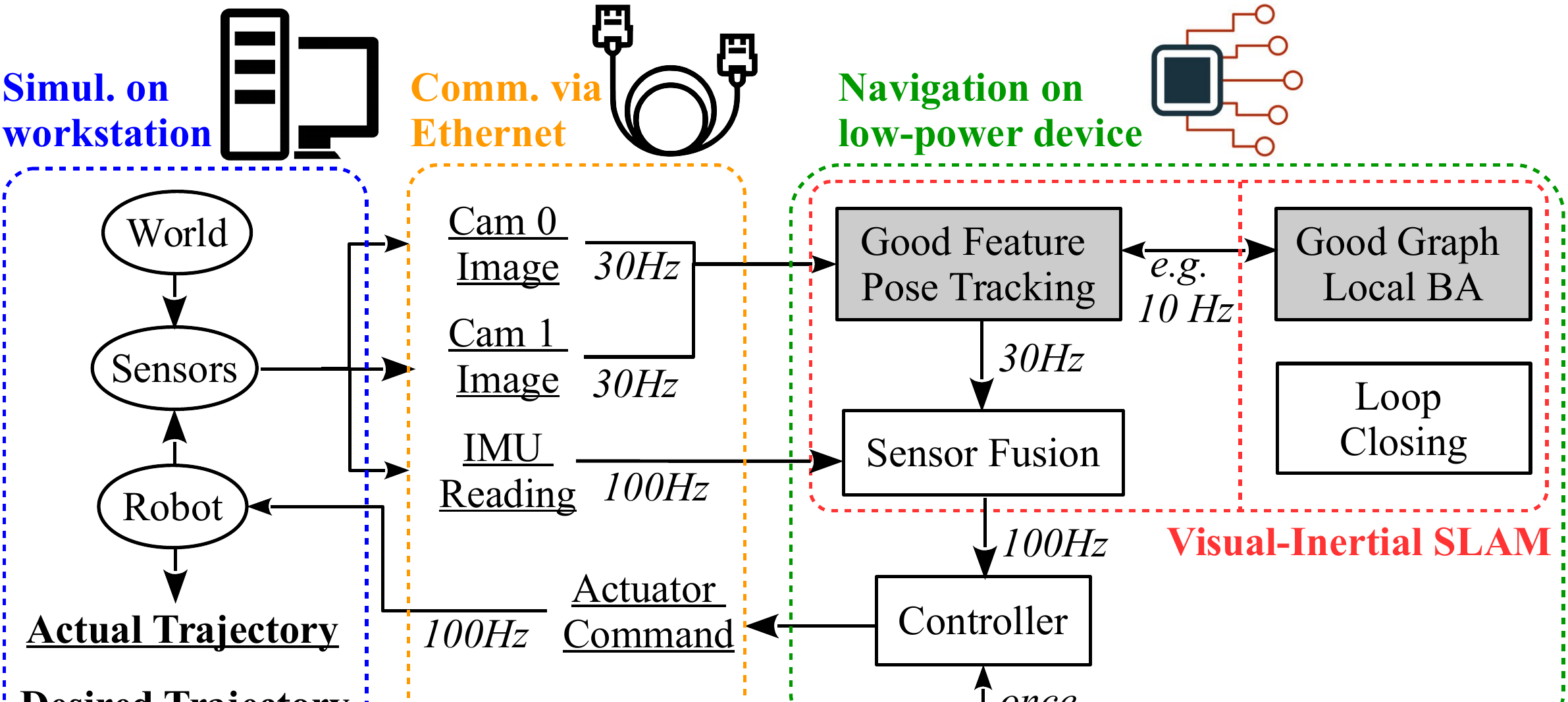}
  \caption{Overview of the closed-loop navigation system, in which 
  {\em GF+GG} is argumented into a loosely-coupled visual-inertial 
  SLAM system with inertial fusion.  Other visual(-inertial) SLAM 
  systems were easily plugged into the navigation system for comparison.
  \label{fig:closedNav_Overview}} 
\end{figure}

The closed-loop navigation benchmarking system introduced in
\cite{zhao2020closednav} is extended to evaluate the proposed Good-Graph
algorithm.  The block diagram in Fig.~\ref{fig:closedNav_Overview}
provides a visual overview of the system.
While the original system \cite{zhao2020closednav} implemented the
simulation and navigation stack on a single workstation, here we
perform a hardware-in-the-loop computing experiment. The processes that
would be performed on a robot are off-loaded to a low-power compute
device, which includes the VI-SLAM, sensor fusion, and trajectory
controller modules. 
The workstation and the low-power device communicate via Ethernet.  
The top row of the figure depicts the two devices and their connectivity.
The bottom row depicts further details of the systems, with the
left-most part representing the ROS/Gazebo simulation and the middle
part the transmitted signals and their rates.  The right-most part
depicts the three major navigation subsystems: 
1) a visual SLAM subsystem taking stereo vision data 
  to generate sparse yet accurate state estimates; 
2) an EKF-based sensor fusion subsystem \cite{lynen2013robust} taking
  both sparse visual estimates and high-rate IMU readings for
  high-rate and accurate positioning; and 
3) a PID position controller \cite{near-identity} taking the high-rate
  output from the sensor fusion subsystem and generating actuator
  commands. 

\subsubsection{Baseline Methods}
The canonical ORB-SLAM ({\em ORB}) \cite{ORBSLAM} and the
front-end-improved {\em GF} variant \cite{zhao2020gfm} serve as
comparison baselines within the same solution class.  Additional
baseline methods using other SLAM approaches include three of the four
visual(-inertial) SLAM systems from the previous standalone VSLAM
evaluation: 
1) filter-based, visual-inertial {\em MSC} \cite{sun2018robust}, 
2) visual-inertial {\em VIF} with sliding window BA \cite{qin2018vins}, and 
3) visual-only {\em SVO} with sliding window BA \cite{SVO2017}. 
{\em ICE} is not included due to the lack of ROS support.

\subsubsection{Simulation Setup}
The simulation setup is a virtual office world of dimensions
20m $\times$ 20m depicted in Fig.~\ref{fig:closedNav_SimuWorld}. 
The virtual world is based on the floor-plan of an actual office, with
texture-mapped surfaces \cite{zhao2020closednav}.
The walls are placed 1m above the ground plane since collision checking
and path planning is outside the scope of this evaluation; we wish to
avoid this as a factor influencing the outcomes.  The differential drive
robot TurtleBot2 \cite{garage2011turtlebot} is commanded to track
reference trajectories within the office world.  Mounted to the robot
are a 30fps stereo camera with an 11cm baseline and an IMU placed at
the base of TurtleBot.  Two commonly-used IMUs are simulated: a high-end
ADIS16448 and a low-end MPU6000.  Data streams from both the stereo
camera and IMU are input to the visual(-inertial) SLAM which then
outputs $SE(3)$ state estimates.  The trajectory tracker
\cite{near-identity} uses the $SE(2)$ subspace of the $SE(3)$ estimate
to track the desired path.

\subsubsection{Experiment Details}

\begin{figure*}[!tb]
  \begin{minipage}{0.49 \textwidth}
  \centering
  \includegraphics[height=1.9in]{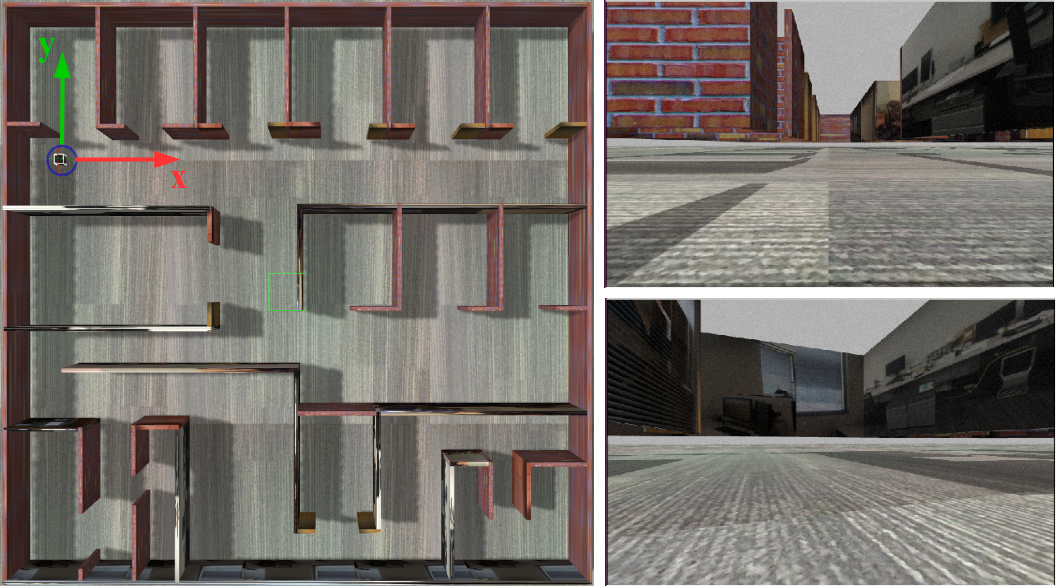} 
  \caption{The virtual office world.
  \textbf{Left: } Top-down view.  The robot starts at the top-left
    corner, facing the long corridor.
  \textbf{Right: } Example images captured by on-board stereo camera 
    (left camera). 
  \label{fig:closedNav_SimuWorld}} 
  \end{minipage}
  \hfill
  \begin{minipage}{0.49 \textwidth}
  \centering
  \includegraphics[height=1.9in]{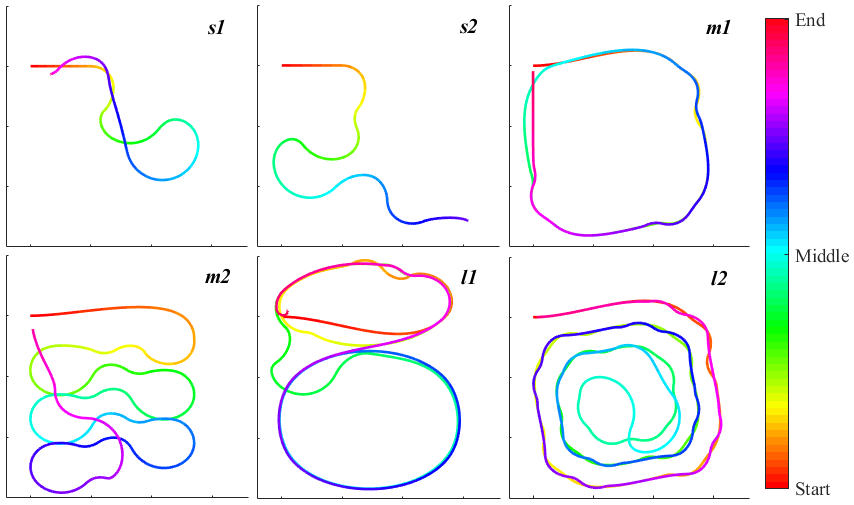} 
  \caption{The 6 desired paths used in closed-loop navigation
    experiments (best viewed in color).  The color-coding indicates
    path progression (see colorbar).
  \label{fig:closedNav_Paths}} 
  \end{minipage}
\end{figure*}

The navigation benchmark consists of six test paths as 
illustrated in Fig.~\ref{fig:closedNav_Paths}, each with different 
characteristics \cite{zhao2020closednav}.
The first two paths are short ($\sim$50m), with few to no re-visits.
The third and forth paths are medium length ($\sim$120m) with many
to few re-visits.  
The last two paths are long ($\sim$240m) with many re-visits.  
All paths start at the world origin where the robot is placed 
(top-left corner of Fig.~\ref{fig:closedNav_SimuWorld}).
Three desired linear velocities are tested: 0.5m/s, 1.0m/s, and 1.5m/s.  
Each closed-loop navigation test configuration (desired path, desired
linear velocity, VI-SLAM method, and IMU) is run 5 times, with the
average performance outcomes presented in the case of no track failures. 
Configurations with at least one track failure are thrown out.

Gazebo simulation and graphics rendering occurs in real-time on an   
Intel Xeon E5-2680 dual-CPU workstation (passmark score 1661 per
thread).  The closed-loop navigation computation includes VSLAM,
visual-inertial fusion, and feedback control.  These processes occur on
a laptop with a low-power Intel Core i7-8550U CPU (single-thread
Passmark score of 2140 and power consumption of 15W).
For reference, most published closed-loop navigation systems 
\cite{scaramuzza2014vision,burri2015real,paschall2017fast,cvivsic2018soft,sun2018robust,lin2018autonomous,oleynikova2020open} 
employ Intel NUC whose CPU scores between 1900-2300 per thread. 
There is a transport delay of approximately 30 ms for the sensor signals
between the simulation workstation and the navigation laptop.

The navigation performance metric is the root-mean-square error (RMSE)
between the desired path and the actual path, averaged over the 5-run
repeats.  
In addition to a 40\% track loss limit, any trajectory with
average RMSE over 10m is considered to be a navigation failure and the
entire 5-run test case is omitted (the dashes).  
The full results of closed-loop navigation discussed here with all 6
visual(-inertial) SLAM systems are also available online \cite{GG_Results}.


\begingroup
\setlength{\tabcolsep}{5pt} 
\begin{table*}[t]
  \small
  \begin{minipage}{0.49\textwidth}
  \centering
  \caption{Navigation Error (RMSE; in m), with High-end IMU ADIS16448
    \label{tab:Nav_RMSE_High}}
  
  \begin{tabular}{|c|c|cccccc|c|}
	\toprule
	\textbf{ } & \textbf{ } & 
	\multicolumn{7}{c|}{\bfseries \small Sequences} \\
	\textbf{Vel.} & \textbf{Meth.} & \textit{s1} & \textit{s2} & \textit{m1} & \textit{m2} & \textit{l1} & \textit{l2} & \textbf{Avg.} \\
	\midrule
	\parbox[t]{1mm}{\multirow{6}{*}{\rotatebox[origin=c]{90}{0.5 m/s}}}
& GF  &\bf 0.12 & 0.12 &\bf 0.18 &\bf 0.12 &\bf 0.22 & - &\bf (0.15) \\ 
&GF+GG& 0.14 &\bf 0.11 & 0.21 & 0.14 & 0.23 &\bf 0.32 & 0.19 \\ 
& ORB &  0.16 & 0.82 & - & 0.32 & - & 5.29 & (1.65) \\ 
& MSC &  0.34 & 0.35 & 0.50 & 0.62 & 0.56 & 0.58 & 0.49 \\ 
& VIF &  - & 0.64 & - & - & - & - & (0.64) \\
& SVO &  2.00 & - & - & - & - & - & (2.00) \\ 
	\midrule
	\parbox[t]{1mm}{\multirow{7}{*}{\rotatebox[origin=c]{90}{1.0 m/s}}}
& GF  & - & 0.12 & 0.24 & 0.20 & 0.42 &\bf 0.36 & (0.27) \\ 
&GF+GG&\bf 0.14 &\bf 0.11 &\bf 0.23 &\bf 0.19 &\bf 0.33 & 0.46 &\bf 0.24 \\
& ORB &  0.24 & 0.17 & 2.40 & - & - & - & (0.94) \\  
& MSC &  0.26 & 0.45 & 0.53 & 0.67 & 0.85 & 0.87 & 0.61 \\ 
& VIF &  5.59 & - & - & 0.88 & - & - & (3.24) \\
& SVO &  1.06 & 8.49 & - & - & - & - & (4.78) \\ 
	\midrule
	\parbox[t]{1mm}{\multirow{7}{*}{\rotatebox[origin=c]{90}{1.5 m/s}}}
& GF  &  0.23 & 0.19 & 0.65 &\bf 0.21 &\bf 0.33 & 0.44 & 0.34 \\ 
&GF+GG&\bf 0.21 &\bf 0.17 &\bf 0.26 & 0.23 & 0.39 &\bf 0.27 &\bf 0.26 \\ 
& ORB &  0.48 & 0.36 & 0.40 & 0.82 & 0.58 & 0.82 & 0.58 \\ 
& MSC &  - & 0.39 & 0.52 & 0.51 & 0.76 & - & (0.55) \\ 
& VIF &  5.40 & 7.80 & 0.58 & - & - & - & (4.59) \\
& SVO &  0.54 & 1.52 & - & - & - & - & (0.88) \\ 
	\bottomrule	
  \end{tabular} 
  \end{minipage}
  \hfill
  \begin{minipage}{0.49\textwidth}
  \centering
  \caption{Navigation Error (RMSE; in m), with Low-end IMU MPU6000
	\label{tab:Nav_RMSE_Low}}

  \begin{tabular}{|c|c|cccccc|c|}
	\toprule
	\textbf{ } & \textbf{ } & 
	\multicolumn{7}{c|}{\bfseries \small Sequences} \\
	\textbf{Vel.} & \textbf{Meth.} & \textit{s1} & \textit{s2} & \textit{m1} & \textit{m2} & \textit{l1} & \textit{l2} & \textbf{Avg.} \\
	\midrule
	\parbox[t]{1mm}{\multirow{7}{*}{\rotatebox[origin=c]{90}{0.5 m/s}}}
& GF  &\bf 0.13 &\bf 0.14 & 1.02 & 0.36 &\bf 0.36 &\bf 0.55 & 0.43 \\ 
&GF+GG& - & 0.17 &\bf 0.36 &\bf 0.35 & 0.63 & 0.66 & 0.43 \\ 
& ORB &  0.25 & 1.18 & - & - & - & - & (0.72) \\ 
& MSC &  0.32 & 0.34 & 0.54 & 0.43 & 0.84 & 0.63 & 0.52 \\ 
& VIF &  - & - & 0.53 & - & - & - & (0.53) \\
& SVO &  0.17 & - & - & - & - & - &\bf (0.17) \\ 
	\midrule
	\parbox[t]{1mm}{\multirow{7}{*}{\rotatebox[origin=c]{90}{1.0 m/s}}}
& GF  &\bf 0.17 &\bf 0.15 &\bf 0.21 &\bf 0.14 & 0.42 &\bf 0.34 &\bf 0.24 \\ 
&GF+GG&\bf 0.17 & 0.19 & 0.26 & 0.48 &\bf 0.39 & 0.49 & 0.33 \\ 
& ORB &  0.22 & 0.18 & 0.44 & - & - & - & (0.28) \\ 
& MSC &  - & 0.41 & 0.43 & 0.39 & 0.56 & 0.62 & (0.48) \\ 
& VIF &  - & - & - & - & - & - & - \\
& SVO &  0.98 & - & - & - & - & - & (0.98) \\ 
	\midrule
	\parbox[t]{1mm}{\multirow{7}{*}{\rotatebox[origin=c]{90}{1.5 m/s}}}
& GF  &  - & 0.25 & 0.63 &\bf 0.17 & 0.44 &\bf 0.40 & (0.38) \\ 
&GF+GG&\bf 0.20 &\bf 0.19 &\bf 0.29 & 0.20 &\bf 0.35 & 0.45 &\bf 0.28 \\
& ORB &  0.43 & 0.43 & 0.36 & 1.02 & - & - & (0.56) \\  
& MSC &  0.31 & 0.36 & 0.59 & 0.51 & 0.66 & - & (0.53) \\ 
& VIF &  - & - & - & - & - & - & - \\
& SVO &  0.79 & 5.32 & - & - & - & - & (3.06) \\ 
	\bottomrule	
  \end{tabular} 
  \end{minipage}
\end{table*}
\endgroup

\subsubsection{Navigation Results on Low-Power Laptop}

Tables~\ref{tab:Nav_RMSE_High} and \ref{tab:Nav_RMSE_Low} provide the
quantitative outcomes for navigation performance with the high-end and
low-end IMUs, respectively.  
Similar to Table~\ref{tab:EuRoC_Stereo_RMSE_Embedded}, the lowest RMSEs 
with the same category (defined by desired path, desired
linear velocity, and IMU) are highlighted in bold.  On 
the rightmost column the average RMSEs are summarized, where the 
average numbers in parentheses indicate more track loss than {\em GF+GG}. 
Across the three speed settings, there were 18 sequences tested per IMU
case. 
%
{\em SVO}, {\em VIF}, and {\em ORB} were the least robust for both IMU cases. 
For the high-end IMU, there were 13, 12, and 5 tracking failures, respectively.
For the low-end IMU, there were 14, 17, and 9 failures.
Prior evaluation has shown SVO to be quite robust
\cite{zhao2020gfm,zhao2020closednav}, however in this case the
data transport delay combined with the typically large pose estimation
error of {\em SVO} undermines stable operation. In what follows, we will focus
on the implementations with better tracking success.

\begin{figure*}[tb]
  	\includegraphics[width=\columnwidth]{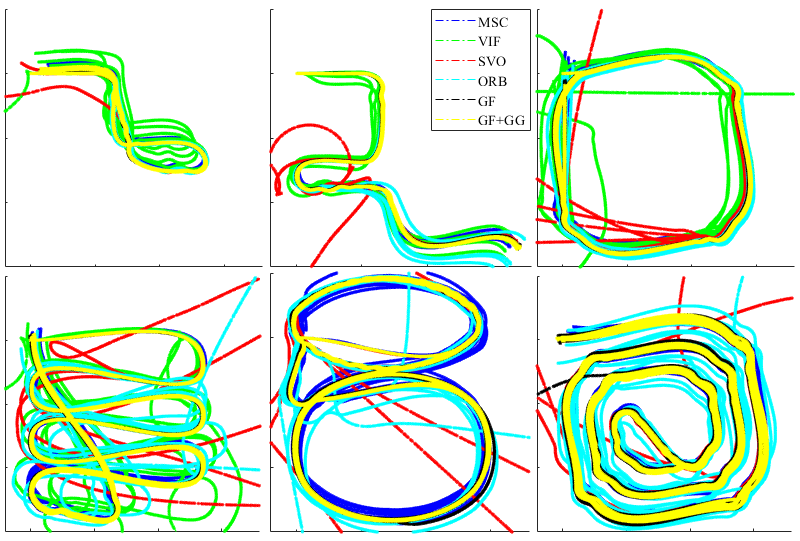} 
  	\hspace{5pt}
    \includegraphics[width=\columnwidth]{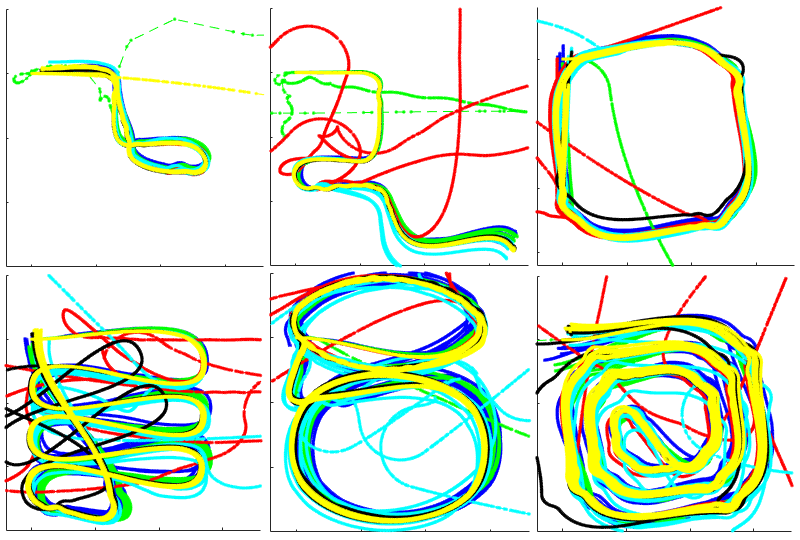}
    \caption{Trajectories the robot traveled for each desired path, 
    with a desired velocity of 0.5m/s.  The trajectories are color-coded by method.
    \textbf{Left}: simulated with a high-end ADIS16448 IMU model.
    \textbf{Right}: simulated with a low-end MPU6000 IMU model.
    \label{fig:Nav_Track}} 
\end{figure*}

Of the remaining methods with low tracking failure counts, the {\em
GF+GG} variant had only 1 track failure over the 36 tests, while {\em
GF} had 3, and {\em MSC} had 4.  Furthermore, {\em GF+GG} had more
consistent error outcomes under the different desired velocities when
compared to {\em GF} and {\em MSC} for both IMU cases. 
For the high-end IMU, the average error of {\em GF+GG} increased
from $0.19$ to $0.26$ RMSE(m), whereas the {\em GF} RMSE went from
$0.15$ to $0.34$, and the {\em MSC} RMSE went from $0.49$ to $0.55$. 
A roughly similar trend for the average RMSE range holds for the low-end
IMU case, though performance is not strictly increasing as a function of
speed, which suggests that the magnitude of the IMU error and the
nominal pose difference influences RMSE outcomes.  Outside of {\em
GF+GG}, the best performing method is {\em GF}.  The inclusion of the
{\em Good Graph} component improves tracking robustness and smooths out
performance variation, though the {\em GF+GG} does lead to a small
increase in RMSE error relative to the best performing statistics for
{\em GF} alone. Across all of the sequences {\em GF+GG} has mean and
standard deviation of $0.285 \pm 0.140$ while {\em GF} has $0.304 \pm 0.197$
(for reference, {\em MSC} statistics are $0.522 \pm 0.162$).
While consolidating all of the data aggregates different scenarios, the
improved outcome consistency is an indicator of algorithm insensitivity
to the nuisance factors associated to the different scenarios (path
type, vehicle speed, and IMU noise level). The insensitivity
demonstrates that adaptively adjusting the SLAM back-end according
to runtime characteristics and solve-time needs does improve overall
performance.

To qualitatively visualize the impact of these different RMSE error
statistics, Fig.~\ref{fig:Nav_Track} depicts the actual trajectories
traveled by the robot with 0.5m/s desired linear velocity, across the
different SLAM implementations. The algorithms are color-coded with all
five runs plotted per case tested. Trajectories leaving the plot bounds
are clipped, thus worst case performance for some methods ({\em SVO},
{\em VIF}) is not completely visualized.
{\em SVO} and {\em VIF} trajectories are completely decorrelated from
the desired trajectories for many of the cases, which is an extreme form
of track loss. {\em ORB} and {\em GF} have similar problems for a couple
of cases. The high track error of {\em MSC} and {\em ORB} is visible for
some of the cases as shifted or warped versions of the target
trajectories, or as lateral offsets of the trajectory versus the
reference one. The traveled and target trajectories do not agree but
are still correlated.
Thicker or multiple spread trajectories within a plot are further
indications of high tracking variance. Fan out of the terminal state is
an indication of pose estimation drift.  Visually, {\em GF+GG} has the
most consistent outcomes of the methods as established by the
quantitative performance statistics.

\section{Conclusion} \label{sec::conc}
This paper describes the {\em Good Graph} modification, which improves the
cost-efficiency of BA-based VSLAM back-ends under computational or
timing limits.  
An efficient algorithm has been developed to select size-reduced graphs
to be optimized in the local BA thread based on condition preservation.  
The desired size of the {\em Good Graph} is determined on-the-fly 
with budget-awareness.  
The {\em Good Graph} algorithm is evaluated in two example scenarios: 
1) VSLAM as a standalone system, and 2) VSLAM 
as a part of closed-loop navigation system. 
The general BA results suggest that the {\em Good Graph} algorithm
preserves the accuracy of 3D reconstruction while reducing the compute
time, or the compute time per point processed.  
VSLAM evaluations are conducted under a variety of computational limits.  
Analysis of the results demonstrate that the {\em Good Graph} algorithm 
successfully addresses compute and time cost awareness for BA-based
back-ends, which improves VSLAM performance when there is performance
loss based on compute or time constraints.
For practical applications such as closed-loop navigation, the {\em Good
Graph} algorithm enables accurate and compute-aware VSLAM, whose
properties improve closed-loop navigation performance. As a general
purpose modification to the BA back-end, \textit{Good Graph} should
apply to other SLAM methods besides ORB-SLAM. Open sourced code
facilitates translation to these implementations \cite{GG_Code}.




%



%

\bibliographystyle{IEEEtran}
\bibliography{./full_references}

%
\begin{IEEEbiography}[{\includegraphics[width=1in,height=1.25in,clip,keepaspectratio]{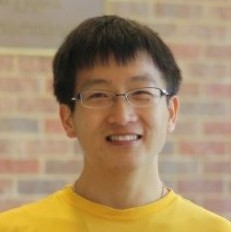}}]{Yipu Zhao}
is a Research Scientist at Facebook Reality Lab (FRL).  Prior to joining FRL, he obtained his Ph.D. in 2019, 
under the supervision of Patricio A. Vela, at the School of Electrical and Computer Engineering, 
Georgia Institute of Technology, USA.  Previously he received his B.Sc. degree in 2010 and M.Sc. degree in 2013,
at the Institute of Artificial Intelligence, Peking University, China.  His research interests include 
visual odometry/SLAM, 3D reconstruction, and multi-object tracking.

Dr. Zhao is a member of IEEE.
\end{IEEEbiography}

\begin{IEEEbiography}[{\includegraphics[width=1in,height=1.25in,clip,keepaspectratio]{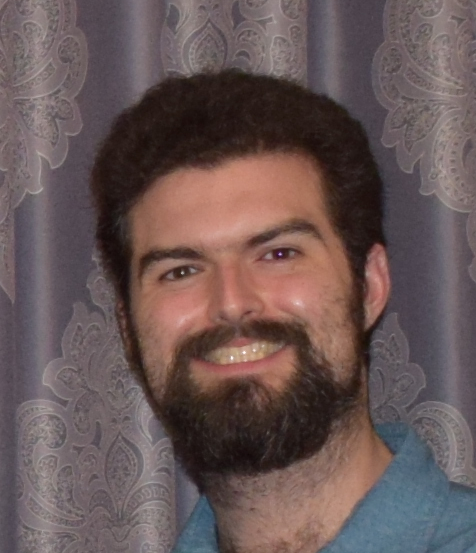}}]{Justin S. Smith} 
is a Ph.D. candidate in the School of Electrical and Computer Engineering, at the Georgia Institute of Technology. 
His research focuses on utilizing perception space and deep learning to improve robot navigation performance, 
especially when under computational resource limit. 
Previously, he earned his B.Sc in Electrical Engineering from Brigham Young University (2012).

Justin S. Smith is a student member of IEEE.
\end{IEEEbiography}

\begin{IEEEbiography}[{\includegraphics[width=1in,height=1.25in,clip,keepaspectratio]{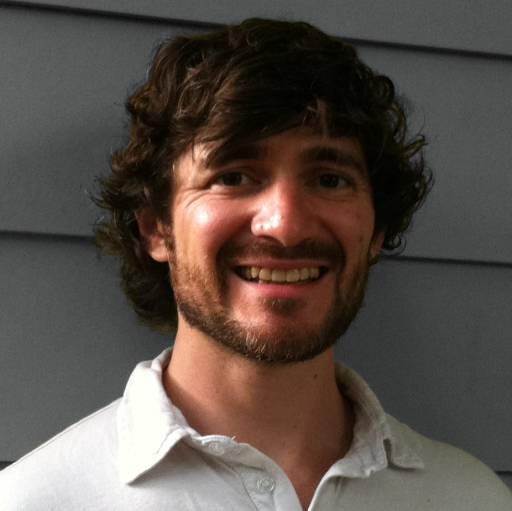}}]{Patricio A. Vela}
is an associate professor in the School of Electrical and Computer Engineering, and the Institute of Robotics and Intelligent Machines, at Georgia Institute of Technology, USA. His research interests lie in the geometric perspectives to control theory and computer vision. Recently, he has been interested in the role that computer vision can play for achieving control-theoretic objectives of (semi-)autonomous systems. His research also covers control of nonlinear systems, typically robotic systems.

Prof. Vela earned his B.Sc. degree in 1998 and his Ph.D. degree in control and dynamical systems in 2003, both from the California Institute of Technology, where he did his graduate research on geometric nonlinear control and robotics. In 2004, Dr. Vela was as a post-doctoral researcher on computer vision with School of ECE, Georgia Tech. He join the ECE faculty at Georgia Tech in 2005.

Prof. Vela is a member of IEEE.
\end{IEEEbiography}

\end{document}